\documentclass[sigconf, 10pt]{acmart}

\usepackage{balance}
\usepackage{soul}
\usepackage{hyperref}
\usepackage{multirow}
\usepackage[utf8]{inputenc}
\usepackage[ruled, lined, longend, linesnumbered]{algorithm2e}
\usepackage{amsmath,multicol}
\usepackage{subcaption,textcomp,comment}

\usepackage{booktabs}
\usepackage{multirow}

\usepackage{paralist}
\usepackage{pifont}

\usepackage{ulem}

\usepackage{array}
\newcolumntype{L}[1]{>{\raggedright\let\newline\\\arraybackslash\hspace{0pt}}m{#1}}
\newcolumntype{C}[1]{>{\centering\let\newline\\\arraybackslash\hspace{0pt}}m{#1}}
\newcolumntype{R}[1]{>{\raggedleft\let\newline\\\arraybackslash\hspace{0pt}}m{#1}}

\newcommand{\mwx}[1]{\textbf{\color{red}{#1}}}

\newcommand{\cdq}[1]{{\color{blue}{#1}}}

\newcommand{\sys}{\texttt{FwdLLM}\xspace}

\setcopyright{none}
\renewcommand\footnotetextcopyrightpermission[1]{} % removes footnote with conference information in first column
\usepackage{wrapfig}
\usepackage{comment}
\usepackage{colortbl}

\usepackage{hyperref}

\usepackage{color,soul}
\usepackage{graphics}
\usepackage{hyperref}
\usepackage{lipsum}
\usepackage{tikz}
\usepackage{enumitem}	% for fine control of list env
\usepackage{listings} % for code listing

\usepackage{booktabs}	% xzl: for fancy tables
\usepackage{lipsum}

\usepackage{capt-of}	% xzl: used to override the ``figure'' and ``table'' in caption.

\usepackage{todonotes}  %xzl: for "missing figures"
\usepackage{subcaption} %xzl: for subfigure. incompatible with package subfig

\definecolor{refkey}{rgb}{0,0,1}
\definecolor{labelkey}{rgb}{0,0,1}

\usepackage{cleveref}
\AtBeginDocument{\DeclareCaptionSubType{lstlisting}}
\crefname{sublstlisting}{listing}{listings}
\Crefname{sublstlisting}{Listing}{Listings}

\usepackage{mathrsfs}	% xzl: for Ralph Smith’s Formal Script Font (rsfs)
\usepackage{amsfonts}

\usepackage[export]{adjustbox} % xzl: flush figure to left/right

\usepackage{boldline}					% jin: use boldline in the table

\usepackage{multirow}

\renewcommand{\paragraph}[1]{\vskip 3pt\noindent\textbf{#1 }}	 % used to be 6pt

  {\begin{list}{$\bullet$}%
     {\setlength{\parsep}{0pt}%
      \setlength{\topsep}{0pt}%
      \setlength{\itemsep}{2pt}}}%
  {\end{list}}
\newcommand\Noted[1]{} % remove highlights.

\definecolor{darkblue}{rgb}{0.0, 0.0, 0.55}
\definecolor{mygreen}{HTML}{ADFF2F}
\definecolor{mylightgray}{gray}{0.8}
  {\begin{itemize}
	[leftmargin=0cm,
		itemindent=.3cm,
		labelwidth=\itemindent,
		labelsep=0pt,
		parsep=0.0pt,
		topsep=0.5pt,
		itemsep=0.5pt,
		align=left]
  }%
  {\end{itemize}}    

  {\begin{enumerate}
	[leftmargin=.cm,itemindent=.5cm,labelwidth=\itemindent,
		labelsep=0pt,
		parsep=1pt,
		topsep=1pt,
		itemsep=3pt,
		align=left]
  }%
  {\end{enumerate}}    

\makeatletter
\def\@copyrightspace{\relax}
\makeatother

\begin{document}

	\title{\sys: Efficient FedLLM using Forward Gradient}
    %\title{Rethinking Cross-Device Federated Learning in the LLM Era}
    % \author{}
    \author{Mengwei Xu}
	% \authornote{Mengwei Xu is the corresponding author.}
	\affiliation{
	\institution{Beijing University of Posts and Telecommunications}
	\country{}
	}

	\author{Dongqi Cai}
	\affiliation{
	\institution{Beijing University of Posts and Telecommunications}
	\country{}
	}

    \author{Yaozong Wu}
	\affiliation{
	\institution{Beijing University of Posts and Telecommunications}
	\country{}
	}

    \author{Xiang Li}
	\affiliation{
	\institution{Beijing University of Posts and Telecommunications}
	\country{}
	}

	\author{Shangguang Wang}
	\affiliation{
	\institution{Beijing University of Posts and Telecommunications}
	\country{}
	}

    \begin{abstract}
Large Language Models (LLMs) are transforming the landscape of mobile intelligence. Federated Learning (FL), a method to preserve user data privacy, is often employed in fine-tuning LLMs to downstream mobile tasks, i.e., FedLLM.
%Though recent efforts have addressed the network issue induced by the vast model size, they have not practically mitigated vital challenges concerning integration with mobile devices, such as significant memory consumption and sluggish model convergence.
A vital challenge of FedLLM is the tension between LLM complexity and resource constraint of mobile devices.

In response to this challenge, this work introduces \sys\footnote{\sys can be accessed at \url{https://github.com/UbiquitousLearning/FwdLLM.git}}, an innovative FL protocol designed to enhance the FedLLM efficiency.
The key idea of \sys is to employ backpropagation (BP)-free training methods, requiring devices only to execute ``perturbed inferences''. Consequently, \sys delivers way better memory efficiency and time efficiency (expedited by mobile NPUs and an expanded array of participant devices).
\sys centers around three key designs:
(1) it combines BP-free training with parameter-efficient training methods, an essential way to scale the approach to the LLM era;
(2) it systematically and adaptively allocates computational loads across devices, striking a careful balance between convergence speed and accuracy;
(3) it discriminatively samples perturbed predictions that are more valuable to model convergence.
Comprehensive experiments illustrate \sys's significant advantages over conventional methods, including up to three orders of magnitude faster convergence and a 14.6$\times$ reduction in memory footprint. Uniquely, \sys paves the way for federated billion-parameter LLMs such as LLaMA on COTS mobile devices -- a feat previously unattained.

\end{abstract}

    \maketitle
    
    % \input{todo}
    % !TeX root = main-Mobisys23-FedPrompt.tex  
\section{Introduction}\label{sec:intro}

Large Language Models (LLMs)\footnote{In this work, we mainly refer LLMs to transformer-based NLP models that exceed 100M parameters.} such as GPTs and LLaMA have showcased an impressive ability to handle generic machine learning tasks~\cite{touvron2023llama}. As foundational models, pre-trained LLMs can be fine-tuned for various downstream tasks and have been applied across a broad range of mobile applications, including but not limited to question answering, personal assistance, and data retrieval~\cite{deeptype, jin2023emsassist, cao2019deqa,sun2020mobilebert, wu2021pecam}. 
Early efforts have been invested to adapt LLMs to mobile devices while maintaining data privacy during the fine-tuning process. Often, these efforts employ federated learning, an approach known as FedLLM~\cite{cai2022autofednlp, zhang2023towards, cai2022federated,cai2022augfedprompt, zhao2023fedprompt,fedbert, wang2023can}.

A salient feature of LLMs is their scalability: by incorporating more parameters, LLMs can continually evolve, achieving higher accuracy or even emergent abilities~\cite{goyal2017accurate, hoffer2017train, you2020limit, sohoni2019low}.
Consequently, contemporary LLMs have grown enormously in size and are hard to be trained even on a GPU cluster~\cite{touvron2023llama}, not to mention mobile devices.
%with GPT-3.5 boasting 175 billion parameters~\cite{gpt35}. Even more compact and mobile-friendly models like LLaMA-7B~\cite{touvron2023llama}
%require a cluster of data center GPUs for training and are too substantial to be accommodated on mobile devices.
%Recent advancements have sought to integrate FedLLM with Parameter-Efficient Fine-Tuning methods (PEFT) like Adapters~\cite{pfeiffer2020adapterhub} and LoRa~\cite{hu2021lora}. By freezing the majority of pre-trained weights during fine-tuning, these techniques have achieved considerable communication reduction between devices and aggregators~\cite{cai2022autofednlp,cai2022federated}.
Recent research of FedLLM~\cite{cai2022autofednlp, zhang2023towards,zhao2023fedprompt,fedbert} primarily addresses the network issue between devices and cloud aggregator, yet the convergence is still lengthy and being impractical for developers.
Through pilot experiments ($\S$\ref{sec:back:exps}), we identify three key obstacles towards practical FedLLM.

$\bullet$ \textbf{Huge memory footprint.}
The predominant on-device training algorithm~\cite{wang2022melon,cai2020tinytl} necessitates extensive memory to store intermediate results such as activations and gradients.
Although fine-tuning could omit most gradients with layers frozen, activations continue to demand considerable memory, often exceeding device capabilities. For example, 3.9 GB is required for RoBERTa-large.
%This memory demand also scales linearly with the batch size and input sequence length~\cite{wang2022melon, yang2020beyond, msl, beltagy2020longformer}, thus constraining LLM scalability.
It results in extra I/O time to swap in/out data~\cite{lee2020fast,peng2020capuchin} and makes the training task a highly likely victim of mobile OS’s low memory killer~\cite{android-lmk}; in either way, the FedLLM convergence is significantly slowed down.

$\bullet$ \textbf{Incompatible with mobile accelerators.}
Mobile SoCs are often furnished with powerful, fast-evolving DNN accelerators (NPUs), e.g., Google Edge TPU and Qualcomm Hexagon that are up to 30$\times$ faster than CPUs.
%can predict with RoBERTa-large within 100 ms -- 15.6$\times$ and 3.8$\times$ faster than its associated 8-core CPU and GPU.
%The capability of NPUs is rapidly evolving as well in recent years (Figure~\ref{fig:motivations-npu-dev}).
%It underscores the opportunity to harness mobile NPUs to expedite FedLLM.
Regrettably, on-device training is unsupported on nearly all mobile NPUs, since they are tailored for inference rather than training, and thus lack the requisite support for training-specific operations like SELECT\_OPS~\cite{selectops} and dynamic gradient updating.

$\bullet$ \textbf{Limited device scalability.}
In FL, only dozens of devices participate in training simtaneously, even when millions of IoT/smartphone devices are available . For instance, Google's deployed FL system samples merely around 1\% of training-ready devices per round~\cite{gboard-fl}, because even a small number of devices can saturate learning performance, meaning additional devices do not further expedite convergence.
%Such a scalability problem is fundamentally rooted at the algorithmic level, specifically within the backpropagation algorithm~\cite{werbos1990backpropagation}, and is thus resistant to resolution from a systems perspective.

%All the aforementioned issues converge to one critical consequence: slow model convergence. For example, fine-tuning a RoBERTa-large model can take tens of hours to converge, as detailed in our experiments. When memory constraints and swapping I/O are considered, this convergence time may increase by an order of magnitude, rendering such costs impractical for most mobile devices or users.
This work leverages a crucial observation: all above issues can be somehow traced to the use of backpropagation-based (termed \textbf{BP}) training algorithm~\cite{werbos1990backpropagation,lecun2015deep} on devices (details in $\S$\ref{sec:back:exps}). This prompts an essential question: \textit{is it feasible to replace backpropagation with a more mobile-friendly training algorithm, thereby reinvigorating the FedLLM protocol?}

\textbf{\sys: training with ``perturbed inferences''.}
We thereby present \sys, the first-of-its-kind system that enables practical and scalable FedLLM through BP-free training algorithm.
Instead of calculating one exact gradient using BP, \sys asks each device to perform perturbed inference:
applying a few self-generated small perturbations to the model weights, and compare how their prediction output deviate from the ground truth labels with the unmodified model.
Intuitively, if a perturbation makes the model more accurate (output closer to labels), the perturbation is likely to direct the model to global optima.
In $\S$\ref{sec:design-forward}, we detail how such intuition leads to a mathmatical form to obtain a BP-free gradient that is an unbiased estimator of the true gradient.
%The deviation and perturbations can be jointly turned into an unbiased estimator of the true gradient.
% devices randomly generate multiple gradient candidates (referred to as perturbations) instead of calculating one exact gradient through backpropagation. Then, they validate these candidates by checking if they closely align with the true labels.
Such BP-free training algorithms~\cite{baydin2022gradients,ren2022scaling,stein1981estimation, liu2020primer, flaxman2004online} have been researched by the ML community for decades but seize very few attentions.
Relying on only inferences, the computation on devices is much more memory efficient and NPU-compatible;
more devices also scale to faster convergence to allow more perturbed inferences simltaneously.

While the idea is intriguing, \sys's design confronts three crucial challenges.
(i) BP-free methods have only shown comparable performance with BP on tiny models like LeNet, as they demand proportionally increased perturbed inferences with its model size~\cite{baydin2022gradients,ren2022scaling, park2023fedfwd}.
%At its current form, it is unlikely to be applied to LLMs.
(ii) How many perturbed inferences are good enough before proceeding to the next round?
It is a vital factor with strong impact on the training convergence.
Using a fixed number perturbed inferences sees up to 3$\times$ longer training delay as compared to the final design of \sys ($\S$\ref{sec:design-adaptive}).
There is no silver-bullet setting that results in the fastest convergence under each condition.
Rather, the optimal setting depends on the specific tasks and models;
it also needs to be adapted on the fly even within the same training session, as the favorable setting drifts over time depending on the model’s learning progress.
(iii) The convergence speed of \sys hinges on the perturbations generated. Contrary to most prior BP-free literature, which randomly samples perturbations from classic distributions~\cite{baydin2022gradients, feng2023does, malladi2023fine}, we discover that this method is often sub-optimal.
Our empirical results in $\S$\ref{sec:design-sampling} shows that most randomly sampled perturbations are of low value (orthogonal to the true gradient) to the convergence.

\sys addresses above challenges with three key designs.

First, \sys integrates the perturbed inferences with parameter-efficient fine-tuning (PEFT) methods like LoRa~\cite{hu2021lora} and Adapter~\cite{pfeiffer2020adapterhub}.
It is based on a crucial observation that the training complexity of BP-free methods scales with its \textit{trainable parameters} instead of total parameters, aligning well with PEFTs that necessitate minimal parameters for fine-tuning.
In fact, the larger the LLM, the fewer PEFT trainable parameters required~\cite{liu2021pre,hu2021lora,he2021towards,pfeiffer2020adapterhub}.
While BP-free training and PEFT are priorly known and have been recently applied to FL in a few literature~\cite{park2023fedfwd,feng2023does,cai2022autofednlp,zhao2023fedprompt}, we are the first to identify their significance in FedLLM and investigate the system implications when they are orchestrated.

\begin{comment}
Conceptually, \sys works alike standard FL protocol:
(1) each device generates $N$ speculative gradients based on a random seed, and applies them to the latest LLM to have $N$ \textit{perturbed} LLMs;
(2) each device obtains $N$ loss differences between each perturbed LLM and the original LLM by executing forward prediction on the local private data;
(3) the cloud aggregates loss differences to estimate the real gradients, and dispatches the new LLM to devices.
%The estimation is based on the forward automatic differentiation (AD) algorithm~\mwx{refs}, which can be traced back to 1960s~\mwx{refs} but is a minority in ML history as compared to backpropagation.
\end{comment}

Second, \sys employs an automatic and systematic strategy to manage the global perturbation inferences for developers.
Unlike traditional FL protocols~\cite{gboard-fl,mcmahan2016federated,li2020federated} that use a static, user-defined metric to control the computing loads on devices,
\sys augments the aggregator with an on-the-fly monitor that controls the timing to aggregate the gradients and proceed to the next round.
Intuitively, as model approaches convergence, BP-free methods need more perturbations to accurately estimate the convergence direction.
\sys leverages a crucial heuristic that the variance across the BP-free gradients uploaded from different devices monotonically increases as model converges, which harmoniously paces with the number of perturbed inferences demanded for fast and stable convergence.
Therefore, \sys proposes a variance-controlled pacing mechanism that the perturbed inferences stops only when the variance observed on aggregator is smaller than a threshold.
\sys also judiciously prioritizes different means (by adjusting participant devices, training data size, and perturbation number) to adapt \texttt{global-PS} to maximize the efficiency.

Third, \sys introduces a discriminative perturbation sampling method that generates perturbations more likely to contribute significantly to convergence.
Concretely, \sys asks devices to bypass the computing of low-value perturbations, i.e., those with nearly orthogonal convergence direction with the true gradients.
To estimate the true gradients that are not known before aggregated on clouds, \sys leverages the opportunity that the gradients direction changes smoothly during FL -- an observation also exploited in prior FL literature~\cite{wang2019adaptive,li2021stragglers}.
Thereby, the server always pre-computes the cosine similarities between the perturbations generated and the computed BP-free gradients of the previous round.
The perturbations with small similarity will be filtered out and not computed by the devices.

We have implemented \sys and evaluated it on 5 typical transformer-based models: ALBERT-base (0.01B), DistilBERT-base (0.07B), BERT-base (0.1B), RoBERTa-large (0.3B), and LLaMA (7B) and 4 classic NLP tasks (both discriminative and generative).
The on-device training is profiled on Google Pixel 7 Pro and Jetson TX2.
%and 3 popular PEFT techniques: Adapter~\cite{pfeiffer2020adapterhub,cai2022autofednlp}, BitFit~\cite{zaken2021bitfit}, and LoRa~\cite{hu2021lora}.
The results demonstrate \sys's impressive performance:
compared to full-model fine-tuning, \sys reduces the training time from 10.9--97.9 hours to 0.2--0.8 hours (up to 217.3$\times$ reduction);
compared to more competitive baselines enhanced by different PEFT methods, \sys still delivers 2.0$\times$--93.4$\times$ speedup (10.6$\times$ on average).
\sys also reduces the memory footprint by up to 14.6$\times$ and 11.5$\times$ compared to them, respectively.
%on average, 488.53$\times$/19.2$\times$ convergence speedup and 7.28$\times$/5.27$\times$ memory footprint reduction over the full finetuning FL baseline and PEFT-enhanced competitive baselines, respectively.
Through orchestration with quantization (INT4), for the first time \sys enables fine-tuning a billion-parameter model like LLaMA over COTS smartphones within only 10 minutes.
The ablation study also shows the significance of \sys's key designs in planning and manipulating perturbations.

\textbf{Contributions}
In this work, we introduce a novel framework that facilitates practical and efficient federated LLM fine-tuning. This is achieved by using BP-free training algorithm. The framework, denoted as \textit{\sys}, innovatively integrates two new techniques that adaptively schedule the number of perturbations to be examined and selectively produce them to enhance their utility. Extensive experiments demonstrate that \textit{\sys} yields substantial improvements over existing baselines. Moreover, \textit{\sys} serves to unify the pathway of on-device inference and training. Rather than viewing these as two separate research domains that necessitate distinct methods and optimizations, \textit{\sys} enables researchers in mobile AI systems and hardware to concentrate on optimizing on-device inference. This focus, in turn, leads to more efficient federated learning processes.

    \section{Background and Motivations}\label{sec:bkgnd}

\subsection{Federated Fine-tuning of LLMs}

Large language model has been an revolutionary technique for its superior performance in serving generic, complex, and few-shot ML tasks.
Training LLMs typically includes two crucial steps:
(i) \textit{pre-training} that endows the models with generic, rich knowledge of images/languages/etc, which requires large amounts of public training datasets and computing devices;
(ii) \textit{fine-tuning} that adapts the pre-trained models for various downstream tasks, which relies on domain-specific, privacy-sensitive data.
Towards a privacy-friendly LLM training pipeline, there is a trend to combine LLM fine-tuning with federated learning, e.g., FedLLM~\cite{cai2022autofednlp, chen2023federated, zhang2023towards}.

We are at very early stage towards practical FedLLM.
The gap between the tight resource constraint of edge devices and the extensive resource demand of on-device LLM training is huge, as demonstrated in both prior studies~\cite{wang2022melon,cai2021towards,cai2020tinytl,xu2022mandheling,xu2018deeptype} and the following experiments.
Recent attempts incorporate parameter-efficient fine-tuning techniques (LoRA~\cite{hu2021lora}, adapters~\cite{pfeiffer2020adapterhub}, and prompt tuning~\cite{liu2021pre}) into FedLLM and see significant improvements in saving the network traffic between devices and aggregator.
Yet they do not fully address many other issues such as excessive memory footprint.

\subsection{Preliminary Experiments of FedLLM}\label{sec:back:exps}

\begin{figure}[t]
	\centering
        \begin{minipage}[b]{0.48\textwidth}
            \includegraphics[width=1\textwidth]{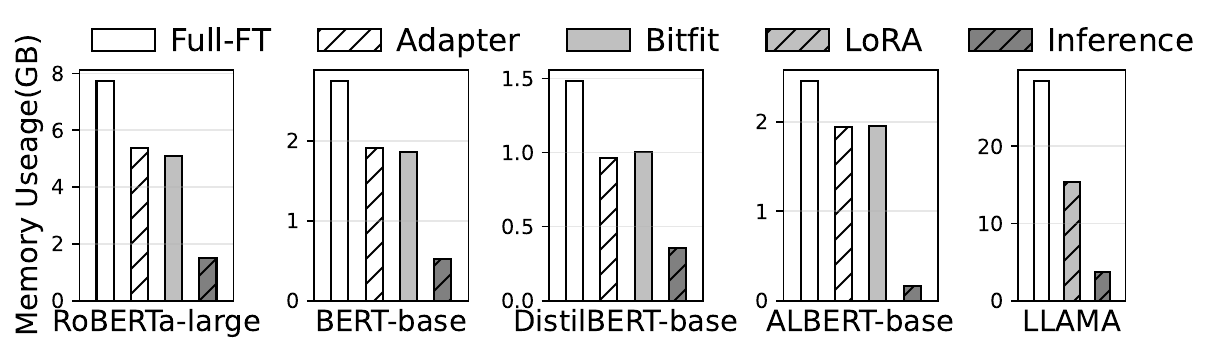}
        \end{minipage}
    \caption{Peak memory footprint of different training methods and inference. Batch size: 8.
    %We only consider the training memory in GPU, i.e., execution memory and input memory are not included.
   % \mwx{Cut llama full training to half (INT4).}
}
	\label{fig:motivations-memory}
\end{figure}

In this subsection, we reveal three crucial issues faced by FedLLM through pilot experiments.

% \usepackage{multirow}
% \usepackage{multirow}
\begin{table}[t]
	\scriptsize
    \includegraphics[width=8.5cm]{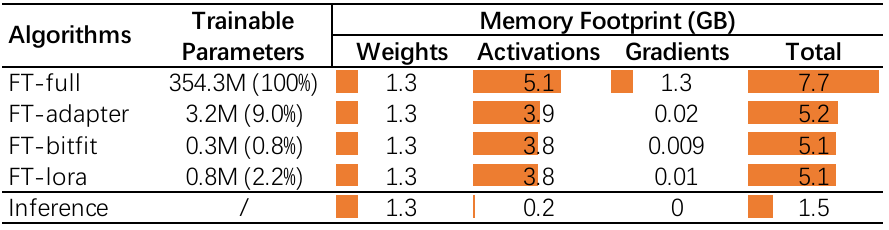}
\caption{The breakdown of memory footprint. Model: RoBERTa-large; batch size: 8. ``FT'': finetuning. 
``Activations'' contain the activations for backward-gradient computation and optimizer states.}
\label{tab:memory-breakdown}
\end{table}

$\bullet$ \textbf{FedLLM is hindered by the memory wall.}
Figure~\ref{fig:motivations-memory} shows the peak memory usage in training various LLMs with a relatively small batch size (8).
The observed memory expense is often unaffordable for edge devices, e.g., more than 7.7GBs for RoBERTa-large and 2.5GBs for ALBERT-base, while the typical mobile devices possess only 4GBs--12GBs DRAM.
More severely, on smartphones, even only a small portion of that memory could be used for training tasks to not compromise user experience~\cite{cai2021towards, lebeck2020end, li2017optimizing}.
In contrast, inference consumes much less memory (e.g., less than 1GB) as it does not need to hold the intermediate computing results in memory as backpropagation does, which linearly scales up with batch size and sequence length.

PEFTs like adapter and bitfit methods cannot fundamentally reduce memory footprint as illustrated in Figure~\ref{fig:motivations-memory}.
They bring only 21.2\%-35.2\% memory savings across different models, which is inadequate to fit certain large models like ROBERTa-large or LLaMA into real mobile devices.
We then break down the memory consumption of ROBERTa-large and summarize the results in Table~\ref{tab:memory-breakdown}.
It explains why reducing trainable parameters cannot bring as significant memory saving: the activations generated during forward pass take up most of the memory usage, which cannot be eliminated even if the weights are not to be updated.

\begin{figure}[t]
	\centering
    \begin{minipage}[b]{0.24\textwidth}
        \includegraphics[width=1\textwidth]{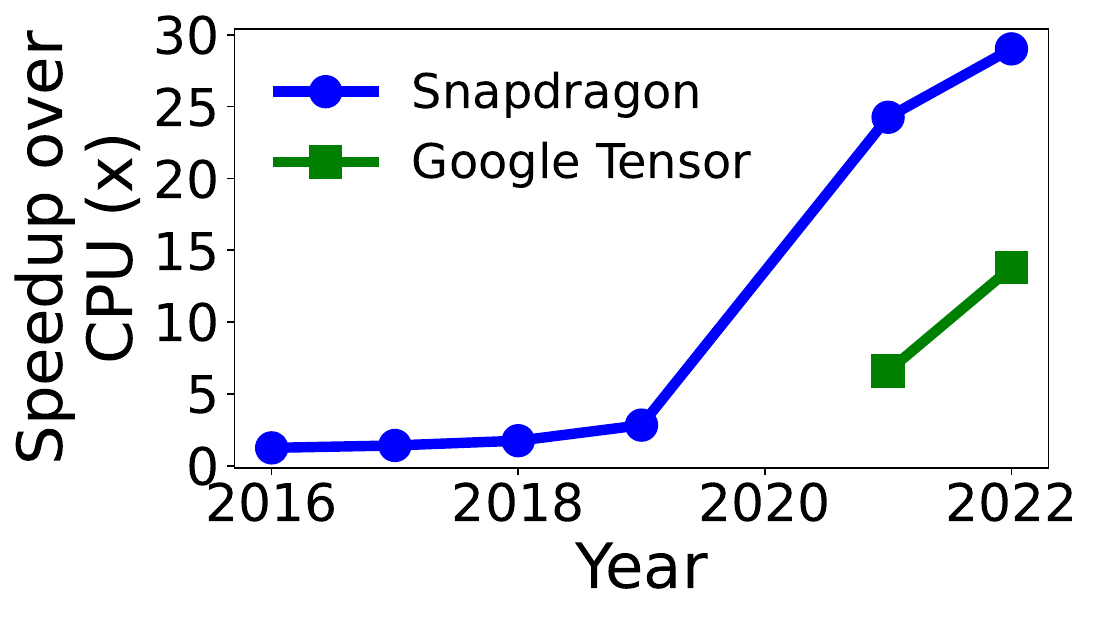}
        \subcaption{ALBERT}
    \end{minipage}~
    \begin{minipage}[b]{0.23\textwidth}
        \includegraphics[width=1\textwidth]{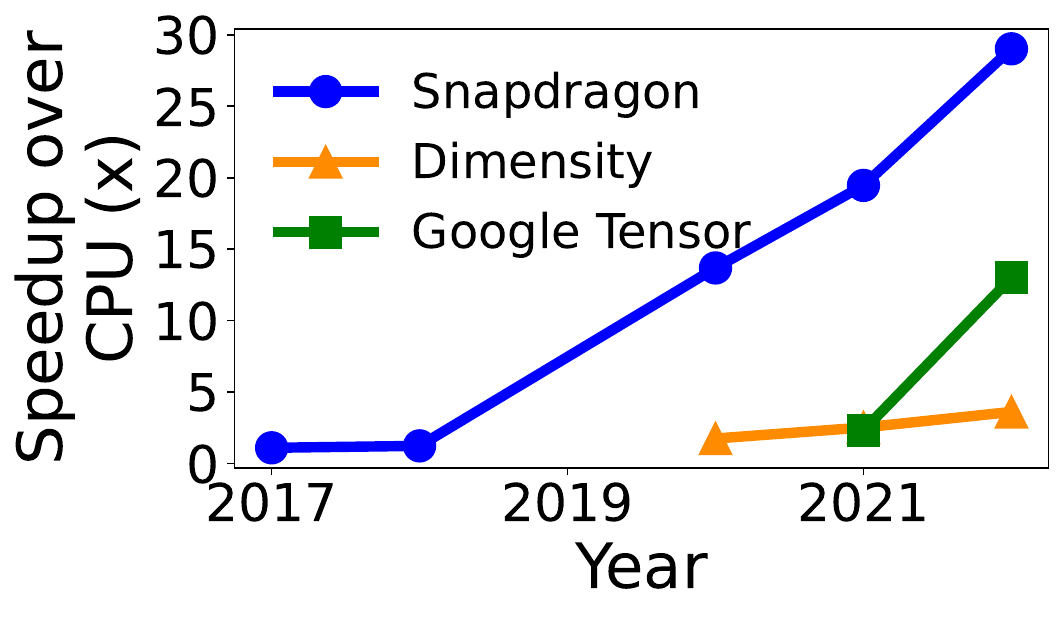}
        \subcaption{MobileBERT}
    \end{minipage}
    \caption{Performance evolvement of mobile NPU. Numbers are from AI Benchmark~\cite{ai-benchmark}.}
	\label{fig:motivations-npu-dev}
\end{figure}

\begin{table}[t]
	\centering
	\resizebox{0.95\columnwidth}{!}{%
		\begin{tabular}{|l|r|rr|r|r|r|}
			\hline
			\multirow{2}{*}{\textbf{Model}} & \multicolumn{3}{c|}{\textbf{Training Time (sec)}}                                      & \multicolumn{3}{c|}{\textbf{Inference Time (sec)}}                                     \\ \cline{2-7} 
			& \textbf{CPU} & \textbf{GPU} & \textbf{NPU} & \textbf{CPU} & \textbf{GPU} & \textbf{NPU} \\ \hline
			%ResNet-50                       & 15.2         & \multicolumn{2}{c|}{}  & 3.0          & 0.3          & 0.08         \\ \cline{1-2} \cline{5-7} 
			ALBERT                          & 17.5         & \multicolumn{2}{c|}{}  & 6.7          & 1.5          & 0.3          \\ \cline{1-2} \cline{5-7} 
			DistilBERT-base                 & 6.9          & \multicolumn{2}{c|}{}  & 3.4          & 0.8          & 0.2          \\ \cline{1-2} \cline{5-7} 
			BERT-base                       & 14.0         & \multicolumn{2}{c|}{N/A}  & 6.9          & 0.8          & 0.3          \\ \cline{1-2} \cline{5-7} 
			RoBERTa-large                   & $^{*}$28.1         & \multicolumn{2}{c|}{}  & 11.7         & 2.9          & 0.8          \\ \cline{1-2} \cline{5-7}
			LLaMA-7B (INT4)          & N/A   & \multicolumn{2}{c|}{}  & 22.1         & N/A          & N/A          \\ \hline
		\end{tabular}%
	}
	\caption{Per-batch (BS=8) training and inference time on Google Pixel 7 Pro. Library: llama.cpp for LLaMA and TFLite for others. ``N/A'': not supported. * is emulated in an infinite memory environment.
		% \mwx{add llama-INT4. its training on CPU is also N/A.}
	}
	\label{tab:motivations-latency}
\end{table}

$\bullet$ \textbf{FedLLM's inability to leverage powerful mobile accelerators.}
Modern mobile devices frequently come equipped with high-end NN accelerators. As Moore's Law approaches its limits, ASIC-based accelerators offer a promising pathway to sustain the growth in device capability in tandem with increasing model complexity. Figure~\ref{fig:motivations-npu-dev} summarizes the speedup achieved by NPUs over CPUs for three popular mobile chip series: Qualcomm Snapdragon (Hexagon), Google Pixels (Tensor TPU), and MediaTek (Dimensity). The findings are compelling: the NPU speedup rate is steadily rising and reaches nearly 30$\times$ on the recent Snapdragon 8+ Gen 1 chip. This underscores the imperative to utilize NPUs for efficient on-device DNN execution.

Nevertheless, on-device training can hardly benefit from mobile NPUs. In Table~\ref{tab:motivations-latency}, our measurements reveal the degree to which NPUs can accelerate DNN inference/training on the Google Pixel 7 Pro. While mobile NPUs significantly reduce inference latency compared to mobile CPUs and GPUs, they offer no support for DNN training. The reason behind this limitation is clear: these NPUs are tailored for inference tasks and thus lack the requisite support for backpropagation-specific operators such as \texttt{BroadcastGradient}, \texttt{ReluGrad}, \texttt{StridedSliceGrad}, and others~\cite{jeong2022band}. Recent work~\cite{xu2022mandheling} has enabled DNN training on Snapdragon Hexagon DSP, but this approach (i) compromises model accuracy due to lower data precision, and (ii) faces scalability challenges with other more proprietary NPUs, such as those found in Google Pixels and Huawei smartphones.

\begin{figure}[t]
	\centering        
        \begin{minipage}[b]{0.48\textwidth}
            \begin{minipage}[b]{0.48\textwidth}
            \includegraphics[width=1.8\textwidth]{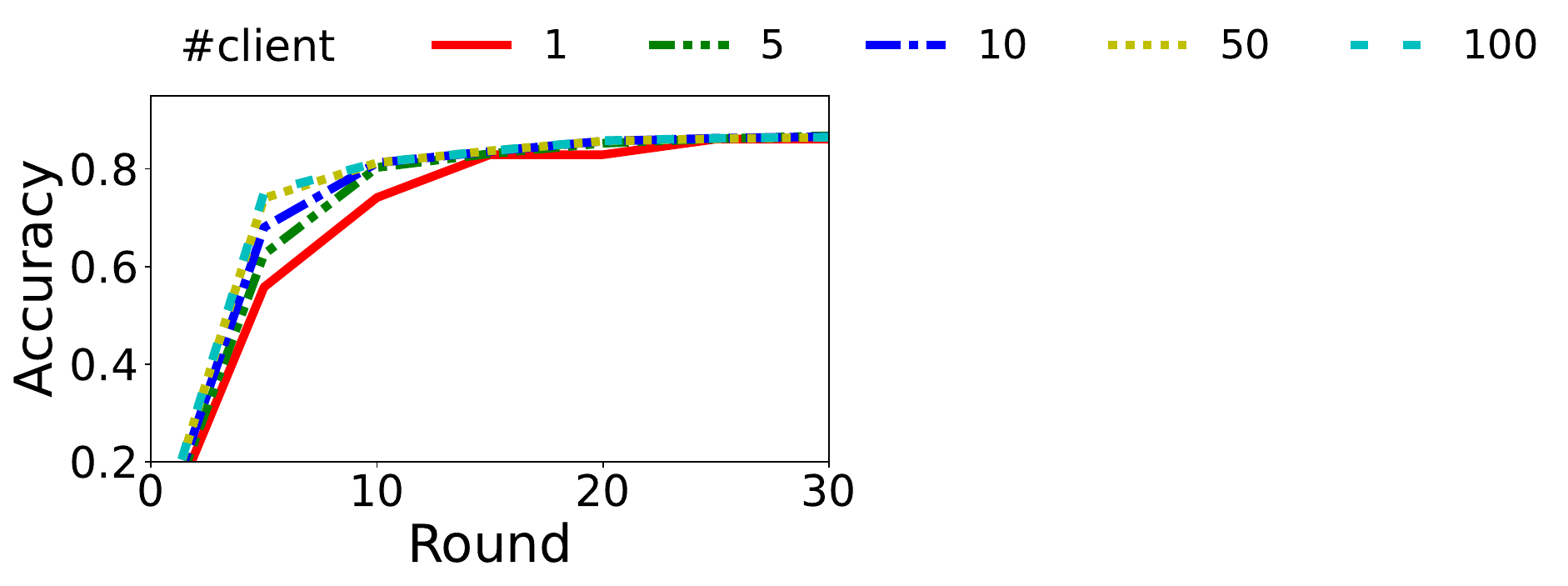}
            \subcaption{Clients (w/ adapter)}
            \end{minipage}~
            \begin{minipage}[b]{0.48\textwidth}
                \includegraphics[width=1\textwidth]{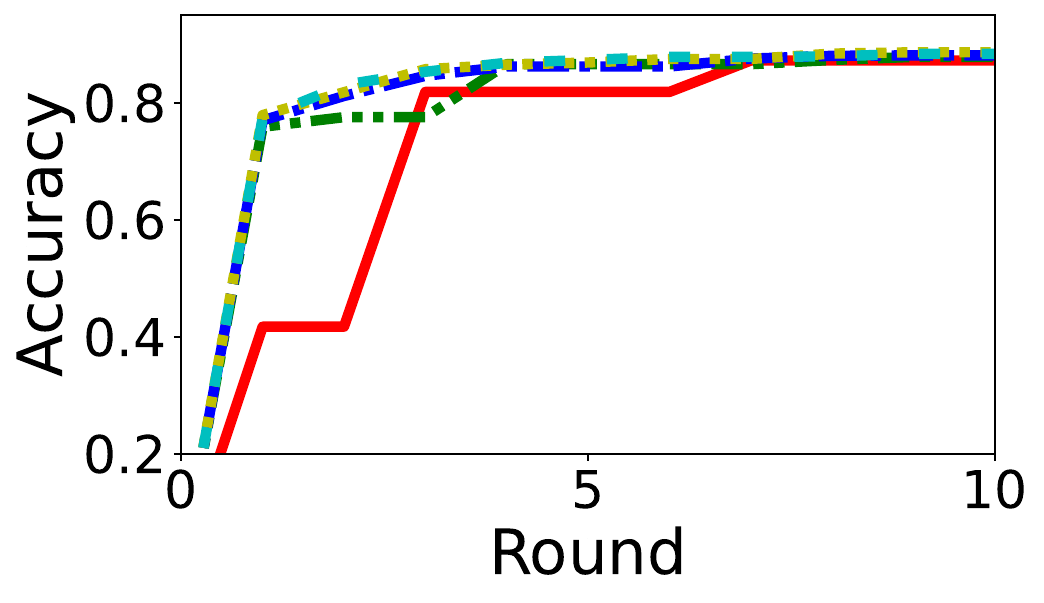}
                \subcaption{Clients (w/o adapter)}
            \end{minipage}
        \end{minipage}
    \caption{Backpropagation-based FL has low device scalability. 
    % \mwx{Why FedSGD? Shall be FedAvg.}
    }
	\label{fig:motivations-scalability}
\end{figure}

$\bullet$ \textbf{FedLLM has low device scalability.}
Backpropagation-based FL struggles to enhance its convergence speed with more participant devices. As depicted in Figure~\ref{fig:motivations-scalability}, involving merely tens of devices per round leads to the saturation of convergence speed, regardless of the utilization of PEFT techniques. Escalating the device count to 100 results in only marginal improvements; for example, it yields just a 1.04$\times$ acceleration to reach an accuracy of 86\% (with adapters).

In real-world scenarios, there could be easily more  than millions devices, such as smartphones and IoTs, that are capable of contributing local data and computing resources. After filtering out devices in non-optimal conditions, such as those with low battery or high utilization, the remaining training-available device number still easily exceeds hundreds. For instance, in Google's deployed FL system~\cite{gboard-fl}, around 10,000 devices are simultaneously available for local training. The failure to scale to a larger number of idle devices seriously constrains the rate at which the model can converge.
As indicated in Table~\ref{tab:motivation-clients}, even when millions of devices are available, the existing literature on FL uniformly adopts the default experimental setting of using no more than 100 devices.
The root cause of this scalability issue can be attributed to backpropagation-based optimizers~\cite{goyal2017accurate, hoffer2017train, you2020limit} and can hardly be addressed at systems aspect.

% \cdq{I think we should claim that we `could' use more clients, but NOT a must.
% Or the reviewer may argue that we could not reach all clients at the same time.}
% \usepackage{graphicx}
\begin{table}[t]
    \centering
    \resizebox{0.95\columnwidth}{!}{%
    \begin{tabular}{|l|l|l|r|r|}
    \hline
    \textbf{Literature}          & \textbf{Venue}    & \textbf{Year} & \textbf{Total Devices} & \textbf{Devices per Round}  \\ \hline
    Hermes~\cite{li2021hermes}    & MobiCom & 2021 & 2,414           & 20 (0.8\%)              \\ \hline
    PyramidFL~\cite{lipyramidfl} & MobiCom & 2022 & 342,477         & 50  (0.01\%)             \\ \hline
    FedAdapter~\cite{cai2022autofednlp}     & MobiCom & 2023 & 1,000           & 15 (0.15\%)              \\ \hline
    FedBalancer~\cite{shin2022fedbalancer} & MobiSys & 2022 & 915 & 100 (10.9\%) \\\hline
    Oort~\cite{lai2020oort}      & OSDI    & 2021 & 1,600,000       & 100  (0.006\%)            \\ \hline
    FedNLP~\cite{lin-etal-2022-fednlp}    & NAACL   & 2022 & 100             & 10   (10\%)            \\ \hline
    C2A~\cite{kim2023client}       & ACL     & 2023 & 100             & 25   (25\%)            \\ \hline
    GradMA~\cite{luo2023gradma}    & CVPR    & 2023 & 100             & 50   (50\%)            \\ \hline
    FedScale~\cite{lai2022fedscale}  & ICML    & 2022 & 1,660,820       & 100  (0.006\%)            \\ \hline
    FjORD~\cite{horvath2021fjord}     & NeurIPS & 2021 & 3,400           & 10   (0.3\%)            \\ \hline
    \end{tabular}%
    }
    \caption{Prior FL literature (mobile/system/AI) use a small ratio of devices in experiments. Maximal numbers are selected if many datasets are used.}
    \label{tab:motivation-clients}
    \end{table}
    \section{\sys Design}\label{sec:design}

\subsection{Overview}
\label{sec:design-overview}
\begin{figure}[t]
	\centering
    \includegraphics[width=0.46\textwidth]{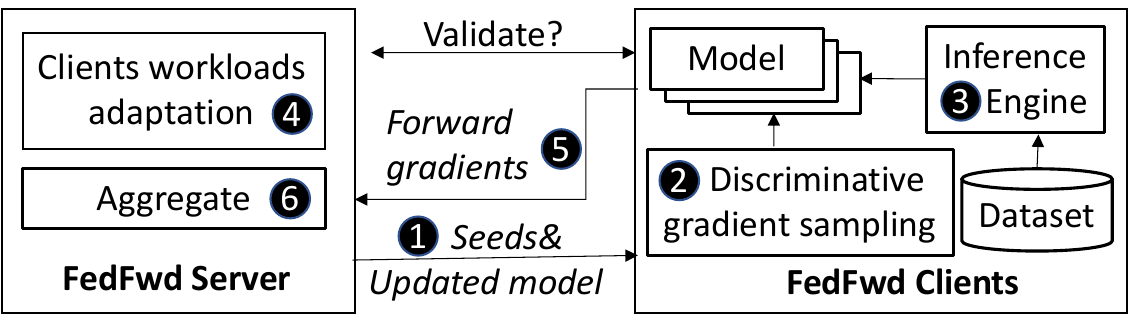}
    \caption{\sys workflow.}
	\label{fig:design-overview}
\end{figure}
\sys is a cloud-device framework that aims to enable practical federated fine-tuning of LLMs across mobile devices.
The key idea is to abandon backpropagation-based gradient descent, but uses ``perturbed inference'' that poses much less memory/compute pressure on devices and can be accelerated by ubiquitous mobile NN accelerators.

\paragraph{Simplified workflow}
As shown in Figure~\ref{fig:design-overview}, \sys employs a similar parameter-server architecture as traditional FedAvg protocol but mainly differs on the local computation.
\ding{182} Per global round, the aggregator first sends the latest LLM updates (only for the trainable weights, denoted as $\mathcal{M}$) and random seeds to each available client. 
\ding{183} Each client discriminatively samples $N$ trainable weights perturbations based on the random seeds, and applies the perturbations to $\mathcal{M}$ to generate $N$ perturbed LLMs (denoted as $\mathcal{M}_{i=1..N}$) ($\S$\ref{sec:design-sampling}).
A perturbation is essentially a vector with the same size as trainable parameter number that is sampled from a uniform distribution.
\ding{184} The client then performs a forward pass on each $\mathcal{M}_{i}$ as well as $\mathcal{M}$ with local training data, from which it gets a forward gradient by comparing their output difference ($\S$\ref{sec:design-forward}).
%The number of perturbations ($N$) and local training data size are critical to the FedLLM performance and is adapted on the fly \ding{185} ($\S$\ref{sec:design-adaptive}).
\ding{185} Forward gradients are validated by the server ($\S$\ref{sec:design-adaptive}) to meet the variance-controlled pace.
Finally, \ding{186} the clients upload the validated forward gradients to the aggregator, where \ding{187} the gradients from different clients are aggregated and applied to $\mathcal{M}$.
The above steps repeat till convergence.

\begin{figure}[t]
	\centering
    \includegraphics[width=0.33\textwidth]{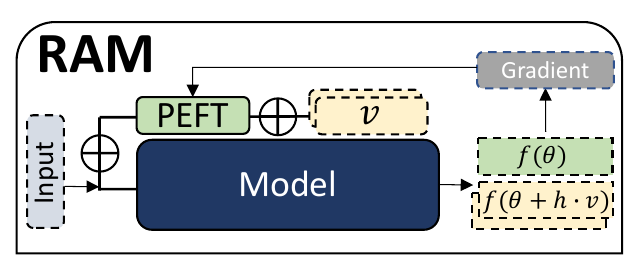}
    \caption{\sys is memory-efficient. 
    % No activation is needed to be stored in RAM for computing gradients.
    Dotted block will be released sequentially after computation.
    %Square blocks indicate the scale while the round blocks indicate the vector.
    }
	\label{fig:design-forward-fl-memory}
	\vspace{-10pt}
\end{figure}
\noindent \textbf{\sys's advantages by design.}
(1) \sys is computationally efficient. For each client, calculating a forward gradient is equivalent to executing inferences twice (i.e., $f(\theta+h \cdot v)$ and $f(\theta)$), making its execution on NPU 19.5--27.5$\times$ faster than computing one backward gradient on a CPU, as previously demonstrated in Table~\ref{tab:motivations-latency}.
(2) \sys is memory-efficient, as it does not require storage of the intermediate activations generated during the forward pass, which contribute to the majority of the memory footprint, as outlined in $\S$\ref{sec:back:exps}. The perturbation weights are parameter-efficient (details in $\S$\ref{sec:design-forward}), and the perturbed LLM can be generated sequentially, with each being immediately released once its inference is completed (Figure~\ref{fig:design-forward-fl-memory}). Thus, the peak memory footprint of \sys can be approximated as $size\_of(\mathcal{M})+2 \cdot trainable\_param$.
(3) \sys is highly scalable with respect to client numbers. As more clients compute forward gradients simultaneously, the aggregated gradient becomes closer to the real gradient, contributing to faster and more accurate convergence. Figure~\ref{fig:design-forward-effectiveness-statistics} illustrates that the continuous addition of perturbations enables the forward gradients to estimate the true gradient with greater precision and stability.
% \input{sec-design-math.tex}

% \mwx{Forward gradient math \& examples: YYY}

\paragraph{Implications on privacy} 
% Clients could integrated with common FL privacy enhancements by doing one step more: computing the forward gradients locally.
% It only adds some small computation overhead to do one multiplication between scale and vector (Equation~\ref{eq:forward-gradient}) and communication overhead to upload vector (parameter-efficient model weights) instead of activations (potentially sensitive) to the cloud.
% Then no activations are uploaded to the cloud, and the cloud only receives the forward gradients.
From the cloud perspective, clients iteratively upload forward gradients, which are unbiased estimations of the true gradients, mirroring traditional FL methods based on backpropagation. Consequently, \sys can be seamlessly integrated with common FL privacy enhancements, including differential privacy~\cite{dwork2008differential,basu2021benchmarking}, secure aggregation~\cite{feng2023does}, and homomorphic encryption~\cite{zhang2020batchcrypt}.
%For instance, to introduce secure aggregation into \sys, we only need a trusted third-party (a common assumption in trust-worthy FL~\cdq{refs})  to negotiate with all clients a group of noise $\left\{\rho_k\right\}_{k=1}^K$ satisfying $\sum_{k=1}^K \rho_k=0$.
%The clients can add the noise $\rho_k$ to the computed forward gradients $g_v(\theta)$ before uploading them to the cloud.
%Thus, server could aggregates those noised gradients correctly without recovering individual gradients.

\paragraph{Unique challenges introduced by BP-free training}
First, BP-free training is not a panacea for all models. 
Existing studies~\cite{baydin2022gradients,ren2022scaling, park2023fedfwd} primarily validate its utility for diminutive models (a few million bytes), which are 1--3 orders of magnitude smaller than standard LLMs that we target.
(2) The quantity of perturbations required to calculate a single forward gradient plays a pivotal role in determining convergence performance (elaborated in $\S$\ref{sec:design-adaptive}). However, this is not trivial to ascertain beforehand and has not been touched in prior literature~\cite{feng2023does, baydin2022gradients,ren2022scaling, park2023fedfwd}.
(3) The convergence speed of BP-free training is hampered by random perturbation generation~\cite{baydin2022gradients,ren2022scaling}.
In following subsections, we will present three novel techniques to tackle those challenges, respectively.

\subsection{Parameter-Efficient BP-Free FedLLM}
\label{sec:design-forward}

%\subsection{Forward-only Federated Learning}
%\label{sec:design-math}

\paragraph{Forward gradient} 
The forward gradient method is selected as our BP-free algorithm because it is based solely on the directional derivative, which can be computed both precisely and efficiently via the forward pass~\cite{baydin2022gradients}.
Formally, to compute the directional derivatives of deep learning functions, denoted as \(f\), with respect to a vector \(v\) at a point \(\theta\), the following equation can be used~\cite{rumelhart1985learning, lecun2015deep}:
\begin{equation}
    \nabla_v f(\theta) = \lim _{h \rightarrow 0} \frac{f(\theta+h \cdot v) - f(\theta)}{h},
    \label{eq:directional-derivative}
\end{equation}
where \(v \in N(0,1)\) represents the weight perturbations, and \(\nabla_v f(\theta)\) is the directional derivative of \(f\) at the model weight point \(\theta\) in the direction \(v\)~\cite{deisenroth2020mathematics}.
In simpler terms, \(\nabla_v f(\theta)\) signifies the slope of \(f\) in the direction of \(v\).
The direction of steepest ascent can be identified using the gradient \(\nabla f(\theta)\).
Nevertheless, determining the gradient \(\nabla f(\theta)\) can be computationally demanding, as it necessitates both a forward pass and a backward pass~\cite{baydin2022gradients}, which we have empirically validated in Section~\ref{sec:bkgnd}.
As an alternative, we leverage the forward gradient \(g_{v}\), which is more cost-effective to compute:
\begin{equation}
    g_v(\theta) := \nabla_v f(\theta) v = (\nabla f(\theta) \cdot v) v,
    \label{eq:forward-gradient}
\end{equation}
where \(g_v(\theta)\) is established as an unbiased estimator of the gradient \(\nabla f(\theta)\)~\cite{baydin2022gradients}.

% Forward gradient allows us to compute $\nabla f(\theta)$ by evaluating $f$ forward $n$ times with direction vectors $v$ taken as standard Gaussian $N(0,1)$.
% \mwx{Compress the above by around 1/3. Give rationales.}

Using Figure~\ref{fig:design-forward-effectiveness-samples} as an illustrative example, we demonstrate the computation of gradients for the function \(z=2(x^2+y^2)\) at the point \((0.5,0.5)\).
We can sample \(n\) direction vectors \(v\) from \(N(0,1)\) and compute \(g_v(\theta)\) for each \(v\). By taking the average of these forward gradients \(g_v(\theta)\), we obtain an estimator for the true (backward) gradient \(\nabla f(\theta)\).
% \mwx{What model? Need more details.}

\begin{figure}[t]
	\centering
    \begin{minipage}[b]{0.48\textwidth}
        \begin{minipage}[b]{0.48\textwidth}
            \hspace*{-3pt}\includegraphics[width=1.15\textwidth]{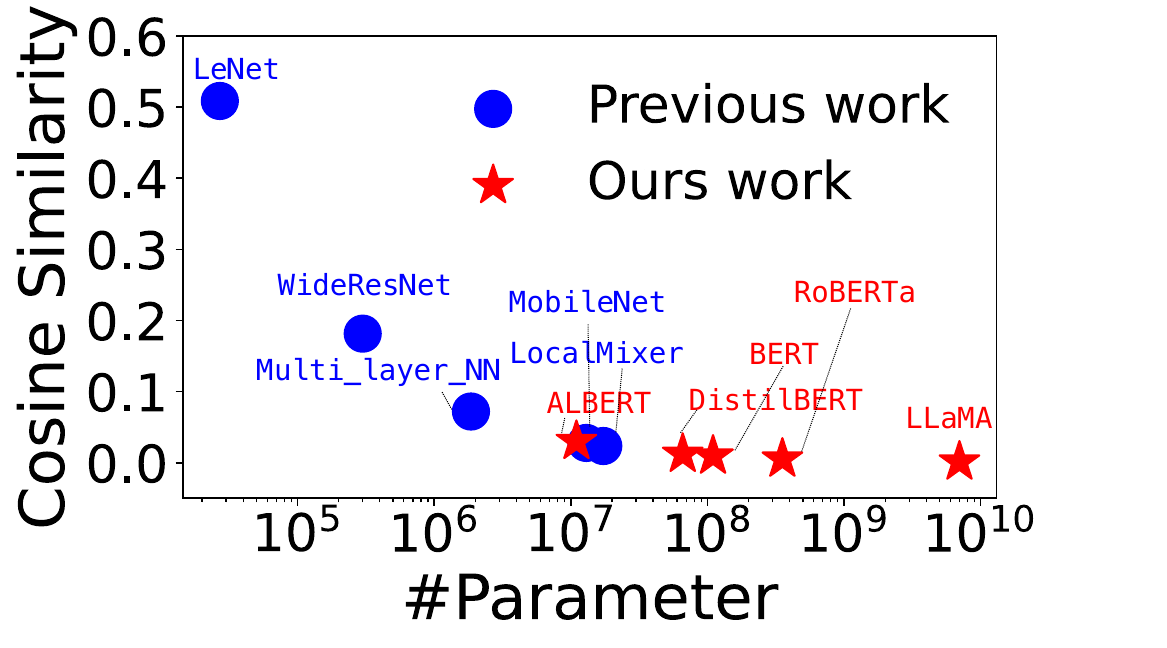}
        \subcaption{Effect of model size.}
        \label{fig:design-forward-effectiveness-samples}
        \end{minipage}~
        \begin{minipage}[b]{0.48\textwidth}
        \includegraphics[width=1\textwidth]{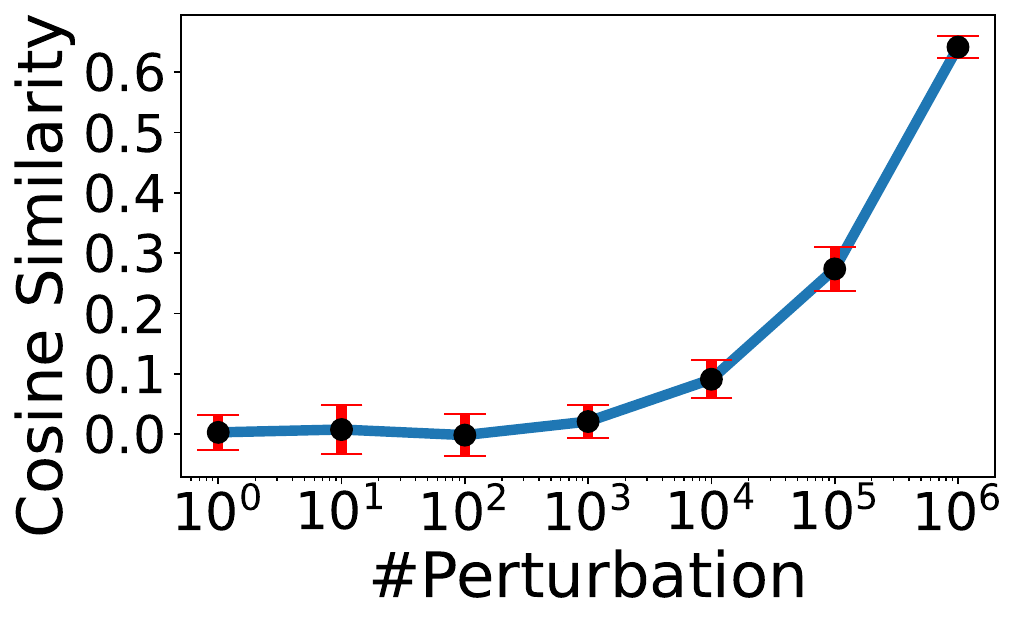}
        \subcaption{Effect of perturbation. 
        % \mwx{Add backpropagation?}
        }
        % Discuss：为什么k越大，越接近真实的gradient，不是均值都是真实梯度嘛？
        % A：因为这里是算得和真实梯度的距离，可以想一下真实梯度是圆心，k越大，圆的半径越小，所以距离越小。
        
        \label{fig:design-forward-effectiveness-statistics}
    \end{minipage}
    \end{minipage}
    \caption{Effectiveness of forward gradients. (a) Increased model parameters make the generated forward gradient unreliable. 
    % for the Beale function \mwx{ref} at x = 1.5, y = -0.1. 
    (b) Adding perturbations could make the forward gradient computed more similar to gold gradients.
    % \cdq{forward-gradients of bert-base layer 0 bias}
    }
	\label{fig:design-forward-effectiveness}
\end{figure}

\paragraph{Forward gradient is not a panacea for all models. }
Although forward gradients have been proposed in previous literature~\cite{barto1987gradient,hinton2022forward,ma2020hsic, baydin2022gradients, sun2022bbtv2,park2023fedfwd}, their practical application has been limited, primarily due to the enormous demands on data and computation.
Specifically, the requirements (i.e., perturbations per batch) grow exponentially with the parameter size.
As shown in Figure~\ref{fig:design-forward-effectiveness}, with the increase of parameter size, the generated forward gradients deviate significantly from the true gradients.
Thus, to obtain reliable forward gradients, the required perturbations per batch increase exponentially with the parameter size.
Prior forward gradient research has largely been restricted to evaluations on small-scale models such as LeNet and WideResNet~\cite{baydin2022gradients,ren2022scaling, park2023fedfwd}.

\paragraph{Parameter-efficient BP-free learning}
To deal with this issue, \sys exploits a key observation that the BP-free training complexity is primarily related to the \textit{trainable} parameter size, rather than the total size.
This observation is intuitively consistent with the mathematical foundation expressed as $f(\theta | \Theta, x)$, where the pre-trained model weight $\Theta$ and input $x$ remain fixed, with only the PEFT weights $\theta$ being tunable.
Fortunately, pre-trained LLMs have accumulated rich generic knowledge, thus requiring only a small number of new parameters to adapt to various downstream tasks.
Therefore, for the first time, \sys integrates BP-free training with PEFT methods.
%This adaptation refers to the process of federated learning with PEFTs, where the larger the LLM, the fewer trainable parameters are needed for each task.
%(2) Mobile devices possess significant scalability, allowing for the simultaneous computation of forward gradients.

%Many off-the-shelf PEFT methods exist, such as LoRA~\cite{hu2021lora}, Adapters~\cite{pfeiffer2020adapterhub}, among others.
%The performance of each PEFT method varies across different models, and new methods continue to emerge~\cite{logan2021cutting,pfeiffer2021adapterfusion,zaken2021bitfit,hu2021lora,he2021towards,liu2021p, sung2022lst,li2021prefix}.
In general,  \sys is compatible with various PEFT methods~\cite{logan2021cutting,pfeiffer2021adapterfusion,zaken2021bitfit,hu2021lora,he2021towards,liu2021p, sung2022lst,li2021prefix} as demonstrated in our extensive offline experiments.
We introduce an offline PEFT profiler designed to automatically identify the most suitable PEFT method for \sys, using public dataset on clouds.
A critical factor influencing \sys performance is the number of trainable parameters.
Consequently, we develope a similarity-aware profile, aiming to identify the optimal PEFT method that maximizes parameter savings while minimizing performance degradation.
Specifically, we train the model using the original parameters and various PEFT methods for a single iteration.
Subsequently, we compute the similarity between the forward gradients and the BP gradients. 
This similarity is then utilized to gauge the efficacy of each PEFT method.
In general, larger LLMs are conducive to more aggressive PEFT methods, like BitFit~\cite{zaken2021bitfit} and LoRa~\cite{hu2021lora}, which yield fewer trainable parameters.
%Our proposed profile is presented in Table~\ref{tab:models}.

\begin{comment}
\paragraph{Computation cost analysis}
According to Equation~\ref{eq:forward-gradient}, the forward gradient $g_{v}(\theta)$ is computed by multiplying the directional derivative $\nabla_v f(\theta)$ with the perturbation vector $v$, where $\nabla_v f(\theta)$ is computed via Equation~\ref{eq:directional-derivative}.
Since cloud is aware of the perturbation vector $v$, the multiplication can be done on cloud.
In conclusion, from client perspective, calculating a forward gradient equals to running two inference, i.e., $f(\theta+h \cdot v)$ and $f(\theta)$.
The time cost of running two inference is \cdq{XX-XX}$\times$ faster than computing one backward-gradients, as we have shown in Table~\ref{tab:motivations-latency}.
\end{comment}

% \mwx{
% have a paragraph to summarize how forward gradient differs from inference, from the computation perspective. For example:    
% From the computing perspective, calculating a forward gradient equals to running inference with a certain batch size and XXX.}

\subsection{Var.-controlled Perturbation Pacing}
\label{sec:design-adaptive}

\paragraph{Key trade-off between accuracy and cost:} In the design of \sys, we identify a crucial trade-off between the rate of model convergence and the computational cost imposed on devices. Specifically, evaluating more perturbations leads to a more accurate forward gradient but also increases the inference cost. We introduce a new metric, global perturbation size (\texttt{global-PS}), defined as the total perturbations 
% \cdq{$\times$ local data size}
% each batch needs one perturbation,
aggregated across all clients per iteration. The success of \sys greatly depends on choosing an appropriate \texttt{global-PS}, a topic that has not been previously explored in studies on forward gradient~\cite{feng2023does,baydin2022gradients,ren2022scaling}.

\begin{figure}[t]
	\centering
    \begin{minipage}[b]{0.48\textwidth}
        \begin{minipage}[b]{0.48\textwidth}
            \includegraphics[width=1\textwidth]{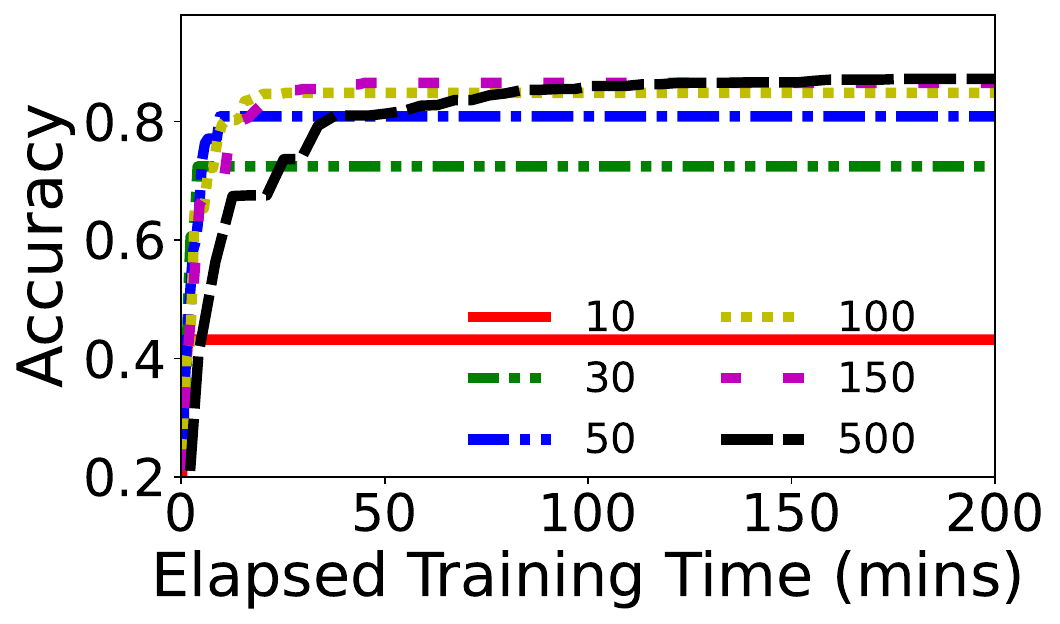}
        \caption{Optimal Global-PS varies across training.}
	    \label{fig:design-planning-configuration}
    \end{minipage}~\hspace{3pt}
        \begin{minipage}[b]{0.48\textwidth}
            \includegraphics[width=1\textwidth]{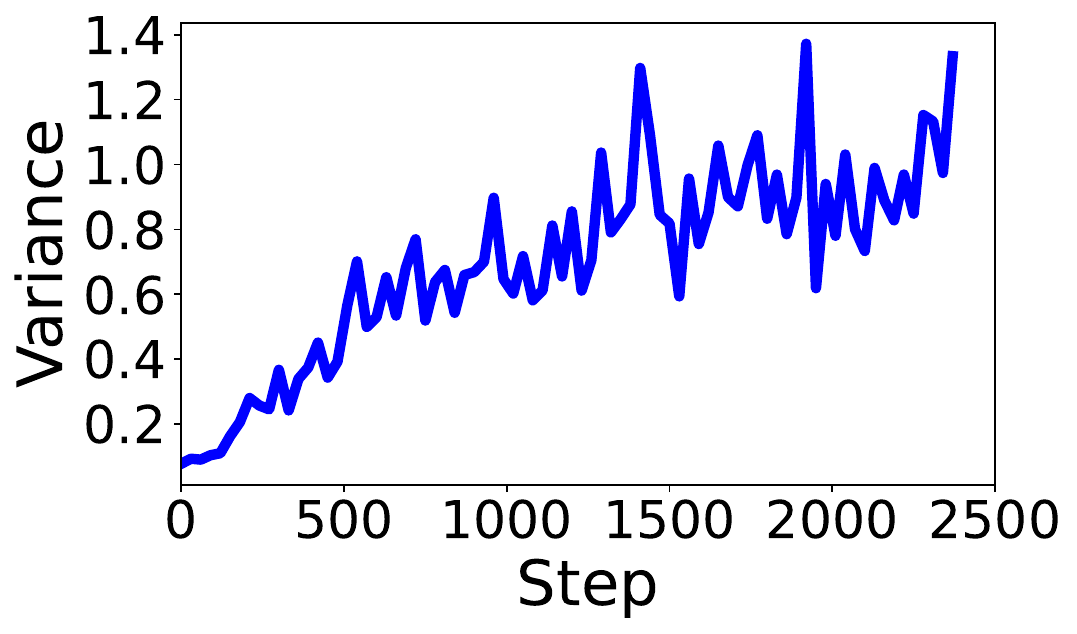}
        \caption{Gradient Variance throught training.}
        \label{fig:design-planning-variance}
    \end{minipage}
    \end{minipage}
\end{figure}
\paragraph{Adaptation of the \texttt{global-PS} on the fly:} There is no universal \texttt{global-PS} setting that can optimize both accuracy and cost across varying scenarios. As illustrated in Figure~\ref{fig:design-planning-configuration}, the optimal \texttt{global-PS} configuration changes throughout the training process, with \sys favoring a monotonically increasing \texttt{global-PS}. For example, in the early training stages of DistilBERT on AGNEWS, the best \texttt{global-PS} is only 3 perturbations per client to reach 80\% relative accuracy to convergence, while achieving 99\% accuracy requires a global-PS of 50, leading to a 16.7$\times$ higher computation cost.

\paragraph{Systematic pacing strategy based on gradient variance:} Requiring developers to manually control \texttt{global-PS} could be complex, as configurations vary across different models and datasets. Instead, \sys offers an automatic and systematic strategy to manage the \texttt{global-PS} parameter, based on the observation that the numerical variance across the forward gradients uploaded by clients increases as the model approaches convergence. 
The variance is defined as $\mathrm{D}(g) = \Vert \frac{1}{2}\left[ \left(\bar{g_{1}}-\bar{g}\right)^2 + \left(\bar{g_{2}}-\bar{g}\right)^2\right]\Vert$,
where \(\bar{g_{1}}\) and \(\bar{g_{2}}\) denote the means of the first and second halves of the forward gradients, respectively, and \(\bar{g}\) denotes the mean of all the forward gradients. 
After 1500 training steps, we observed that this variance escalated from 0.078 to 1.182, marking a 15.2$\times$ increase. 
Such increased variance may necessitate more perturbations to accurately estimate the real gradient. 
\sys simply uses a predefined variance threshold that must be met before aggregating gradients across devices, with the threshold being the only hyperparameter to tune. 
An empirically selected range of 0.1--0.5 has been observed to perform well across various models and datasets.

\paragraph{Prioritizing methods to adapt \texttt{global-PS}:} The \texttt{global-PS} can be increased through three methods: (1) involving more devices; (2) having each device test more perturbations.
% \cdq{(3) having each device test each perturbation on more local data.} 
\sys first prioritizes adding more devices, as concurrent computation facilitates fast convergence. Once device availability reaches its maximum, \sys turns to the second method, asking clients to test more perturbations, for two reasons: (i) perturbations can be quickly and infinitely generated, and (ii) the result of the original LLM ($f(x)$) can be reused when calculating the directional derivatives on multiple perturbations ($f(x+v)$ and $f(x)$), thereby reducing the required forward propagations from $2*N$ to $N+1$.

\paragraph{Validating-computing pipeline:}
\sys meticulously designs a pacing pipeline to validate incoming gradients.
The primary principle is to ensure uninterrupted local forward gradient computations. 
After transmitting the forward gradients to the cloud aggregator, clients proceed to compute the forward gradients for the succeeding perturbation. 
Upon accumulating sufficient forward gradients to surpass the variance threshold, the aggregator instructs the client to halt forward gradient computations. 
It then aggregates the received forward gradients and dispatches the updated model to the client. 
This streamlined process guarantees timely reception of up-to-date pacing information by the server, eliminating futile waiting times for client feedback.
Note that the validation step on cloud aggregator is lightweight, e.g., less than 10ms in our experiment setup.

\subsection{Discriminative Perturbation Sampling}
\label{sec:design-sampling}

\begin{figure}[t]
    \centering
    \begin{minipage}[b]{0.48\textwidth}
    \begin{minipage}[b]{0.48\textwidth}
        \hspace*{-3pt}\includegraphics[width=1.05\textwidth]{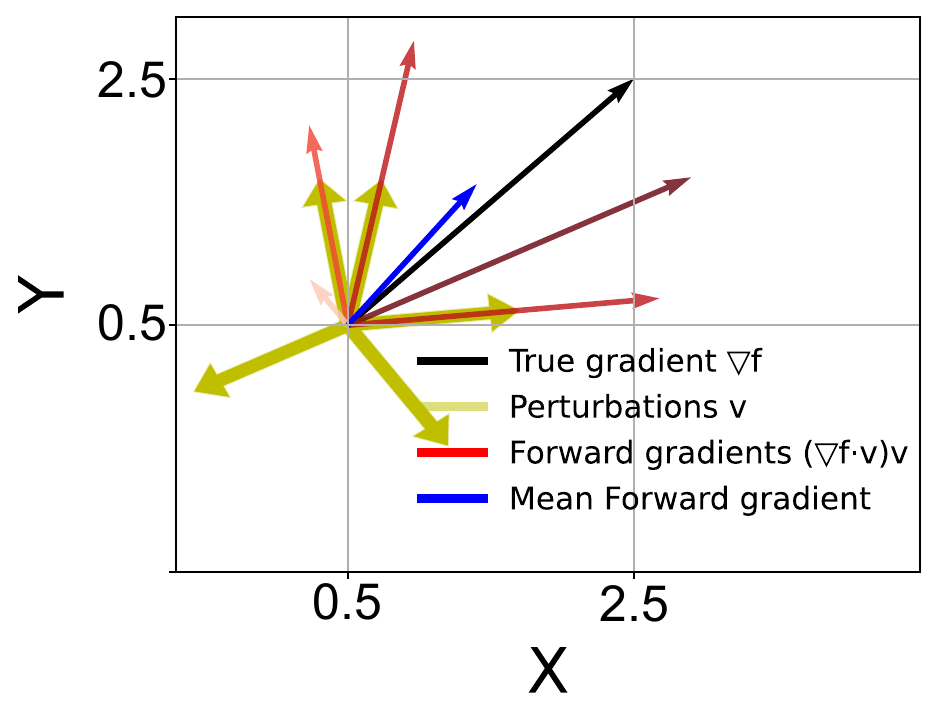}
    \subcaption{Samples}
    \label{fig:design-forward-effectiveness-samples}
\end{minipage}~
\begin{minipage}[b]{0.48\textwidth}
    \includegraphics[width=1\textwidth]{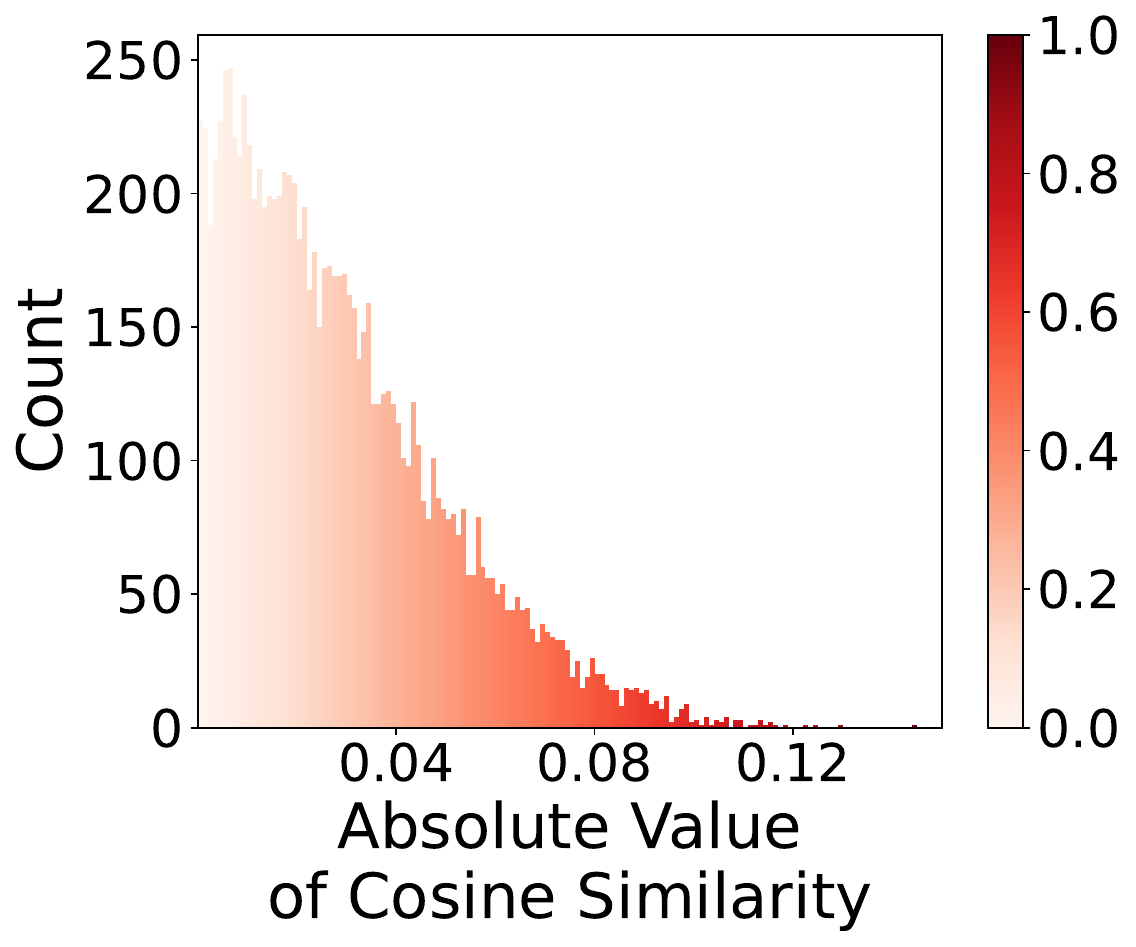}
\subcaption{Statistics}
\label{fig:design-sampling-statistic}
\end{minipage}
\end{minipage}
    \caption{
        % Discriminative gradients sampling design rationale. 
    Most of the gradients are nearly orthogonal to target gradients thus contributing little. 
    % \mwx{(b) randomly generated 768 matrics}
    }
	\label{fig:design-sampling}
\end{figure}

The convergence speed of \sys primarily hinges on the perturbations generated before training. Contrary to most prior forward-gradient literature, which randomly samples perturbations from classic distributions such as Gaussian functions~\cite{baydin2022gradients, feng2023does, malladi2023fine}, we discover that this method is often sub-optimal. \sys instead employs a discriminative approach, selectively sampling perturbations that are \textit{more likely to contribute larger gradients to the model's convergence}.

\paragraph{Minimal contribution from most forward gradients:} Specifically, two randomly generated vectors in high dimensions could be orthogonal~\cite{donoho2009observed, vershynin2018high, tropp2011improved}. Recall that our forward gradient is calculated via $g_v(\theta):=\nabla_v f(\theta) v$, where $\nabla_v f(\theta)=\nabla f(\theta) \cdot v$ is the dot product of the gradient vector $\nabla f(\theta)$ and the perturbation vector $v$. Figure~\ref{fig:design-sampling-statistic} shows our analysis of 10,000 perturbations generated while training DistilBERT on AGNEWS. We find that the majority have cosine similarity with the gradient vector $\nabla f(\theta)$ near zero; specifically, over 60\% have a similarity less than 0.03, and they collectively contribute less than 29.6\% of the final forward gradients. These perturbations contribute minimal forward gradient to model convergence.

\paragraph{Similarity-aware discriminative sampling:}
We propose a strategy to manipulate the perturbation generation process, making it more conducive to convergence. 
Figure~\ref{fig:design-forward-effectiveness-samples} illustrates that when perturbation $v$ is near orthogonal to gradient $\nabla f(\theta)$, the resulting forward gradients $g_v(\theta)$ (shallower red) contribute less.  
Based on this observation, we can select perturbations exhibiting high cosine similarity with the gradient vector $\nabla f(\theta)$.
This technique eliminates the need to compute negligible contributions from certain perturbations. 
Our methodology leverages the opportunity that the gradients direction changes smoothly during FL -- an observation also exploited in prior FL literature~\cite{wang2019adaptive,li2021stragglers}. 
Therefore, it's sufficient to compute the cosine similarities only between the perturbations and the forward gradients from the preceding round.
For practical implementation, perturbation manipulation computations are outsourced to the cloud. 
The cloud first generates a set of random seeds, filtering out perturbations with low cosine similarities to the prior round's forward gradients. 
The resultant seeds are then dispatched to clients for local perturbation generation.

    \section{Implementation and Setups}\label{sec:impl}

\paragraph{\sys prototype}
We have fully implemented the \sys prototype atop FedNLP~\cite{lin-etal-2022-fednlp}, one of the state-of-the-art frameworks to develop and evaluate FL methods on NLP tasks.
There are two primary ways to implement forward gradients: numerical differentiation and analytical differentiation.
We use the former as it is almost identical to forward inference.
We therefore implement it with functorch library~\cite{functorch}.
% to make parameter perturbation more convenient.
For various PEFT methods:
adapter is implemented with AdapterHub~\cite{pfeiffer2020adapterhub}, a library that facilitates the integration of different pre-trained adapters for downstream tasks;
LoRa and BitFit are implemented by ourselves.
The quantized LLaMA training is based on AutoGPTQ library~\cite{autogptq}.
%use adapterhub to add Adapter and LoRA, and freeze all layers except the bias layer to achieve BitFit.
% Directional derivatives are computed by the \cdq{which pip wheel?} function in PyTorch.
% \mwx{how about lora? other libs?}

\begin{table}[t]
	\resizebox{\columnwidth}{!}{%
	\begin{tabular}{l|l|r|r|l}
		\hline
		\textbf{Models} & \textbf{Arch.} & \textbf{Params.} & \textbf{PEFT} & \textbf{Infer. Libs} \\ \hline
		ALBERT-base~\cite{lan2019albert}     & Encoder-only   & 12M                   & BitFit              &      TFLite~\cite{tflite}        \\ 
		DistilBERT-base~\cite{sanh2019distilbert} &  Encoder-only              & 66M                   &      Adapter         &      TFLite~\cite{tflite}            \\ 
		BERT-base~\cite{devlin2018bert}       &  Encoder-only              & 110M                  &      Bitfit         &       TFLite~\cite{tflite}           \\ 
		RoBERTa-large~\cite{liu2019roberta}    &  Encoder-only              & 340M                  &      Bitfit         &        TFLite~\cite{tflite}         \\ 
		LLaMA~\cite{touvron2023llama}        &  Decoder-only              & 7B                    & LoRA             &    llama.cpp~\cite{llamacpp}           \\ \hline
	\end{tabular}
}
\caption{Tested models using PyTorch on TX2.}
\label{tab:models}
\vspace{-10pt}
\end{table}

\paragraph{Models}
We evaluate \sys mainly on five popular LLMs, as shown in Table~\ref{tab:models}.
Four of them are BERT-like models based on transformer encoders:
(1) ALBERT-base (12M)~\cite{lan2019albert};
(2) DistilBERT-base (66M)~\cite{sanh2019distilbert};
(3) BERT-base (110M)~\cite{devlin2018bert};
(4) RoBERT-large (340M)~\cite{liu2019roberta}.
Those four models are extensively used in prior FedNLP research~\cite{cai2023FeS,lin-etal-2022-fednlp,cai2022autofednlp}.
Apart from that, we also evaluate \sys on the SoTA open-sourced generative language model
(5) LLaMA-7B (INT4)~\cite{touvron2023llama}.
To hold the whole LLaMA model in memory, we quantize it to INT4 format using GPTQ~\cite{frantar2022gptq}.
As far as we know, \sys is the first attempt to apply federated learning to a billion-size models.
%Above models have all been supported by Huggingface Transformer~\cite{wolf2019huggingface}.
\sys selects different PEFT methods for different models with its offline profiler presented in $\S$\ref{sec:design-forward}:
Adapter~\cite{pfeiffer2020adapterhub} for DistilBERT;
LoRA~\cite{hu2021lora} for LLaMA;
and BitFit~\cite{zaken2021bitfit} for others.
%The selection is mainly empirically determined based on their trainable parameter size under different PEFT methods.
%because current experiments show that larger models need more aggressive PEFT methods and 
%The rationale behind this is that involving too many trainable parameters will make forward gradient training unstable.
%And BitFit is the most aggressive one because it only tunes bias of each parameter in the network, i.e., only \cdq{0.1\%} of the total parameters.
The pre-trained weights of above models are from Huggingface~\cite{wolf2019huggingface}.

\paragraph{Datasets}
We experiment with four popular NLP datasets:
(1) \texttt{AGNEWS}~\cite{zhang2015character} is a news classification dataset with 4 classes.
The number of training samples for each class is 30K and testing 1.9K.
(2) \texttt{YAHOO}~\cite{zhang2015character} is a topic classification dataset with 10 categories.
Each category contains 140K training samples and 5,000 testing samples.
(3) YELP-Polarity (\texttt{YELP-P})~\cite{zhang2015character} predicts a polarity label
%by considering stars 1 and 2 negative, and 3 and 4 positive 
based on restaurant reviews.
The number of each polarity is 280K/19K for training/testing.
(4) SQuAD-v1.1 (\texttt{SQUAD})~\cite{rajpurkar2016squad} is a commonly used version of stanford question answering dataset.
By default, we uniformly divide the datasets into 10 for Squad, 1,000 clients for AGNEWS and YELP-P, 10,000 clients for YAHOO.
% ~\mwx{why only 1000 for YELP-P?}.
For non-iid settings, we follow prior literatures~\cite{lin-etal-2022-fednlp,cai2022autofednlp} to divide datasets into skewed label distribution.

\paragraph{Hardware}
As prior FL literature~\cite{lai2020oort,lipyramidfl,li2021hermes,lin-etal-2022-fednlp}, our experiments are carried out in an semi-emulation manner on two GPU servers each with 8 $\times$ NVIDIA A100.
The on-device training time is obtained on two popular edge devices:
(1) Google Pixel 7 Pro (Pixel) is a popular mobile phones that is equipped with a Google Tensor G2 TPU, a Mali-G710 GPU and a Octa-core CPU.
It runs Android 13 OS.
We use TFLite~\cite{tflite} to train the four BERT-variant models and llamap.cpp~\cite{llamacpp} to run inference with LLaMA.
The inference and training speed is previously illustrated in Table~\ref{tab:motivations-latency}.
%However, constrained by the OS, it is not able to train different models via GPU in the near way as server.
(2) Jetson TX2 (TX2)~\cite{tx2} is a widely used edge board equipped with a 256-core NVIDIA Pascal GPU and a  Dual-Core NVIDIA Denver 2 64-Bit CPU.
%It runs linux OS.
%so as to be able to training different models via GPU in the near way as server.
We use PyTorch~\cite{pytorch} for on-device inference and training.
Note that Jetson TX2 has no NPU and is not the primary target platform of \sys.
%, therefore we involve it as supplemental result.

\paragraph{Metrics}
We mainly report time-to-accuracy metric and on-device runtime cost (memory, network, energy). 
The target convergence accuracy is 0.88 for AGNEWS, 0.65 for YAHOO and 0.82 for YELP-P, the same as prior work~\cite{lin-etal-2022-fednlp,cai2022autofednlp,zhang2015character}
%Besides, we also report the resource cost imposed in an FL process, including the total amount of energy consumed on data transmitting and training computation on each client; the total amount of network traffic; and the peak memory usage.

\paragraph{Baselines}
We compare \sys to the following baselines:
(1) \texttt{Full-FT} always fine-tunes the whole model~\cite{lin-etal-2022-fednlp}.
(2) Adapter tuning (\texttt{Adapter}) introduces a small tunable module between transformer layers and freeze other parameters~\cite{pfeiffer2020adapterhub,houlsby2019parameter}.
(3) Bias tuning (\texttt{BitFit}) only tunes bias of each layer in the LLM~\cite{zaken2021bitfit}.
(4) \texttt{FedAdapter} is the SoTA FedNLP fine-tuning framework that incorporates adapter tuning with layer freezing techniques~\cite{lin-etal-2022-fednlp}, along with a progressive training paradigm to identify the optimal adapter configuration automatically~\cite{cai2022autofednlp}.
All baselines use BP-based training; thus, they are limited to using smartphone CPUs, as previously discussed in Section \ref{sec:back:exps}. In contrast, \sys can also utilize mobile GPUs and NPUs.
A recent work~\cite{xu2022mandheling} ports training to mobile NPU, yet does not support transformer models.

\paragraph{FL settings}
Unless otherwise stated, \sys and all baselines select 100 clients per round following prior work~\cite{lin-etal-2022-fednlp,cai2022autofednlp, horvath2021fjord,li2021hermes}.
% This is because, as previously shown in $\S$\ref{sec:back:exps}, 10 clients already saturate the convergence performance of backpropgation-based FL methods; while \sys is able to scale its convergence speed with more clients.
% Notably, not every device is training ready due to the runtime and battery status~\cite{gboard-fl}; yet 1,000 clients are commonly available in a popular mobile app.
% We will also compare the performance of \sys and baselines under different training-ready client numbers.
\sys and all baselines use the same set of hyper-parameters as prior work~\cite{lin-etal-2022-fednlp,cai2022autofednlp}: local training epoch as 1; mini-batch size as 8; learning rate as 0.01; max sequence length as 64 for AGNEWS and 256 for others.
The default FL aggregating algorithm is FedSGD~\cite{mcmahan2016federated} as later experiments will show that FedAVG~\cite{gboard-fl} underperforms FedSGD due to the asymmetric compute/network cost.
Network bandwidth is set to 10Mbps by default as prior literature~\cite{lin-etal-2022-fednlp,cai2022autofednlp,cai2023FeS}.
    \section{Evaluation} \label{sec:eval}

%In this section, we first present the overall performance of \sys in Section~\ref{sec:eval-e2e}. 
%We demonstrate \sys's scalability with large generative LLMs in Section~\ref{sec:eval-llama}. 
%Section~\ref{sec:eval-cost} assesses the system cost, while the effectiveness of individual components is evaluated in Section~\ref{sec:eval-ablation}.

\subsection{Convergence Performance on Sub-billion-sized Models} \label{sec:eval-e2e}
\label{sec:eval-e2e}

\begin{figure*}[t]
    \centering
    \begin{minipage}[b]{1\textwidth}
        \begin{minipage}[b]{0.24\textwidth}
            \includegraphics[width=3.55\textwidth]{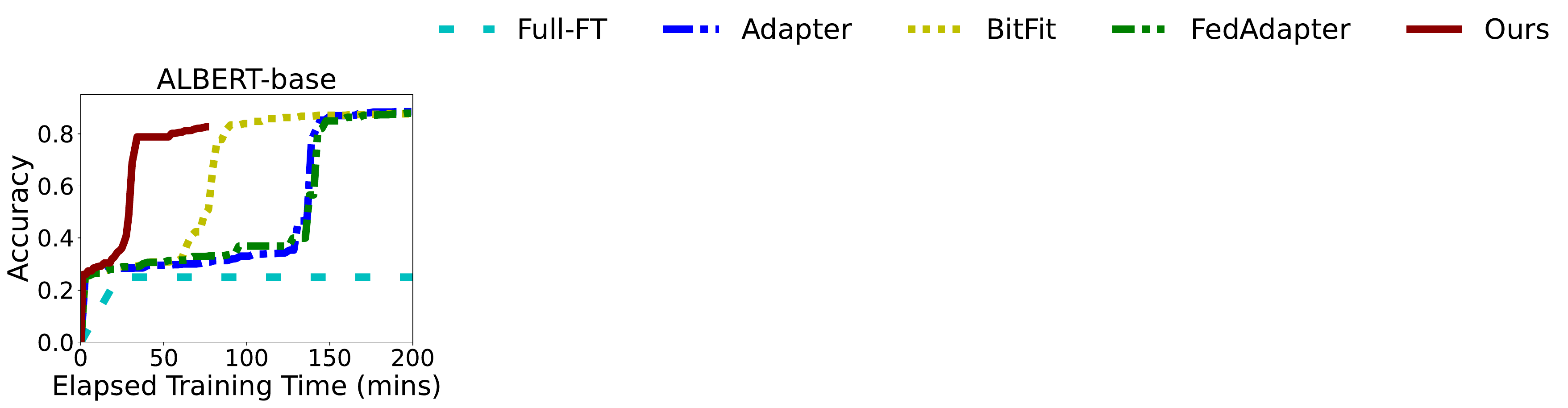}
        \end{minipage}
        ~
        \begin{minipage}[b]{0.24\textwidth}
            \includegraphics[width=1\textwidth]{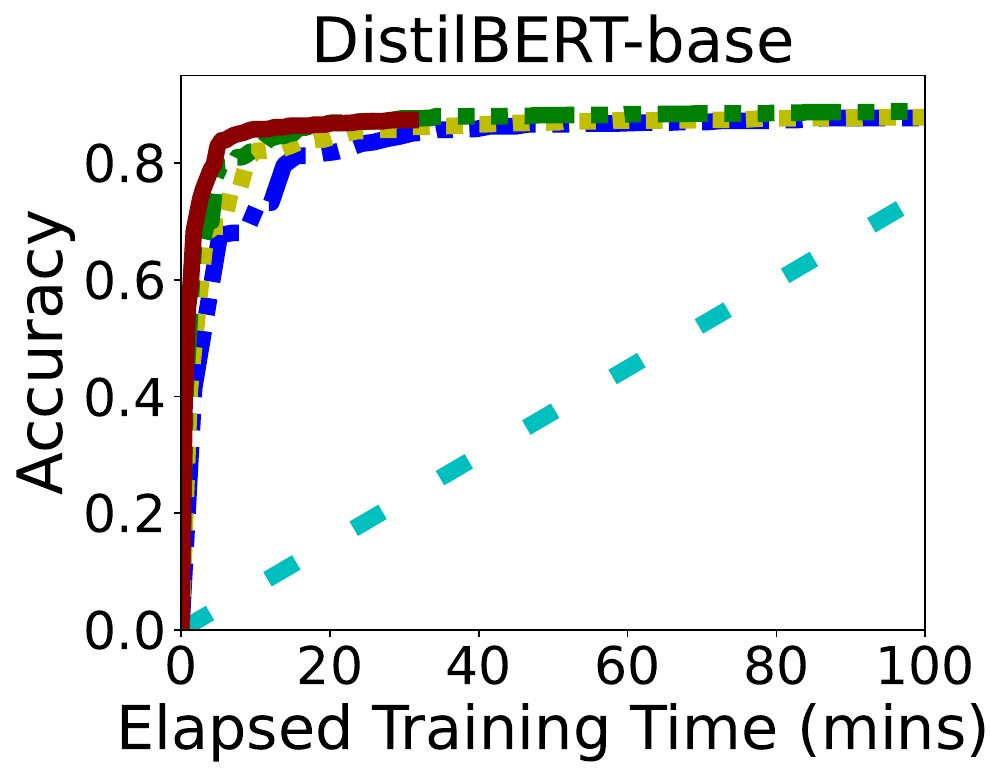}
        \end{minipage}
        ~
        \begin{minipage}[b]{0.24\textwidth}
            \includegraphics[width=1\textwidth]{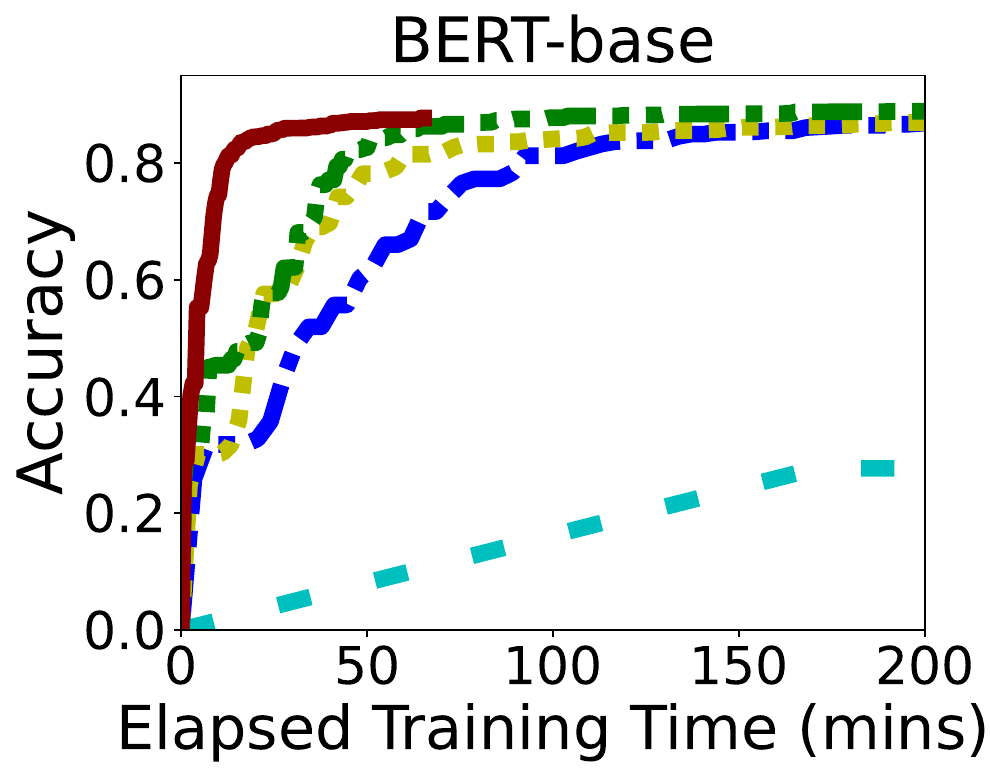}
        \end{minipage}
        ~
        \begin{minipage}[b]{0.24\textwidth}
            \includegraphics[width=1\textwidth]{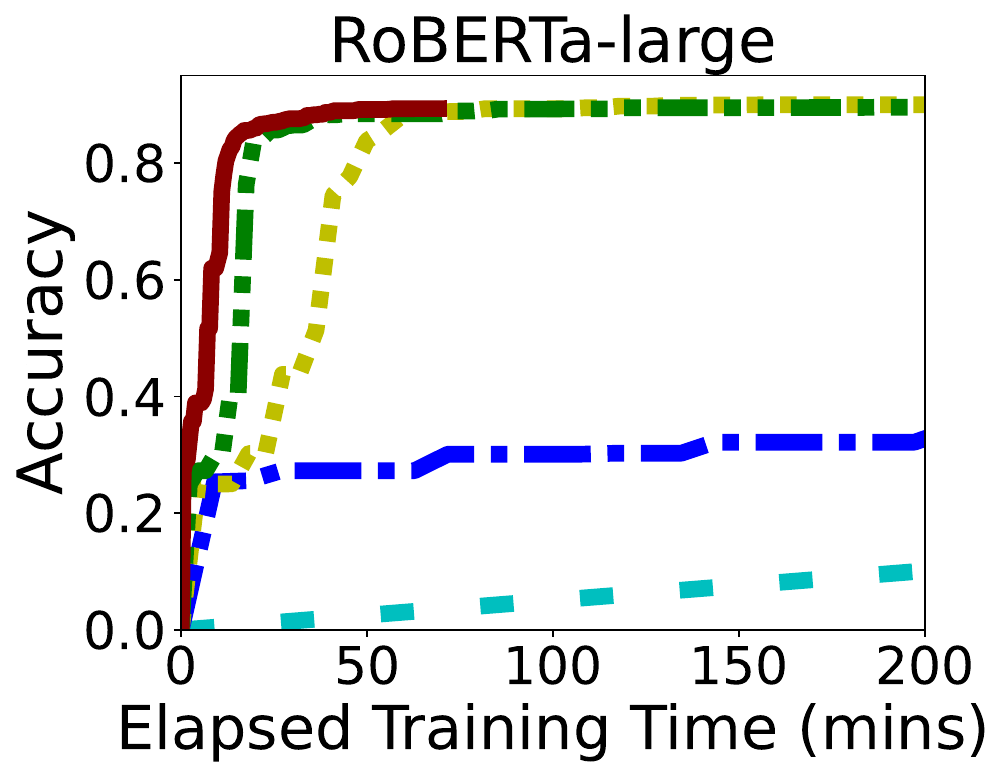}
        \end{minipage}\subcaption{\texttt{AGNEWS}}
    \end{minipage}
    
    \begin{minipage}[b]{1\textwidth}
        \begin{minipage}[b]{0.24\textwidth}
            \includegraphics[width=1\textwidth]{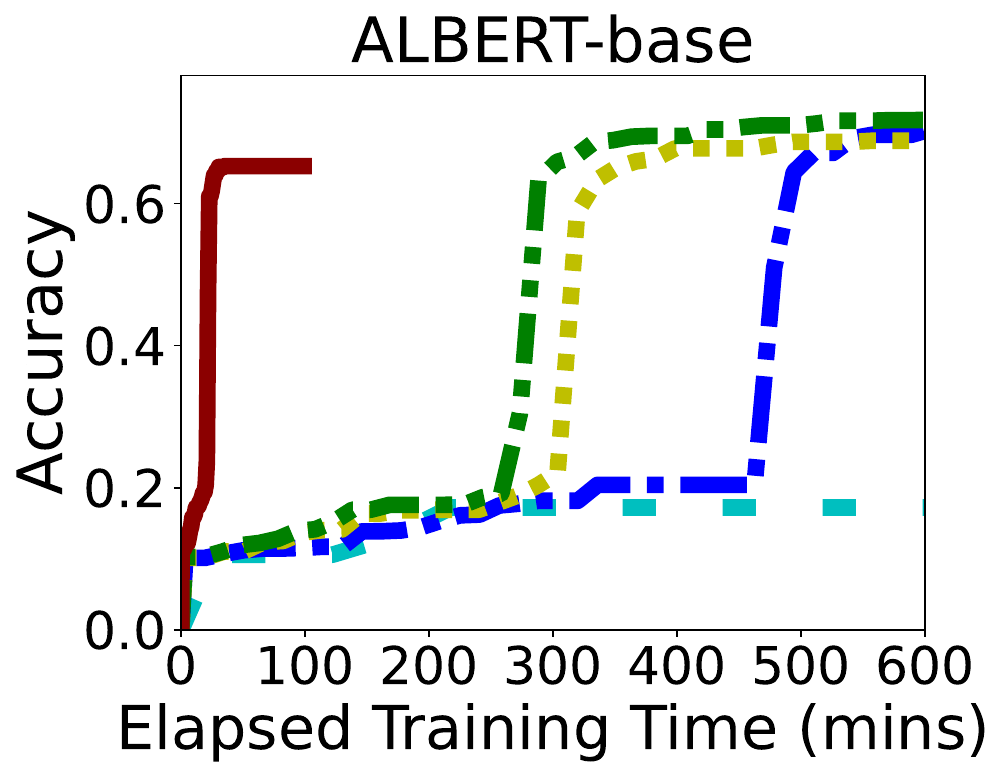}
        \end{minipage}
        ~
        \begin{minipage}[b]{0.24\textwidth}
            \includegraphics[width=1\textwidth]{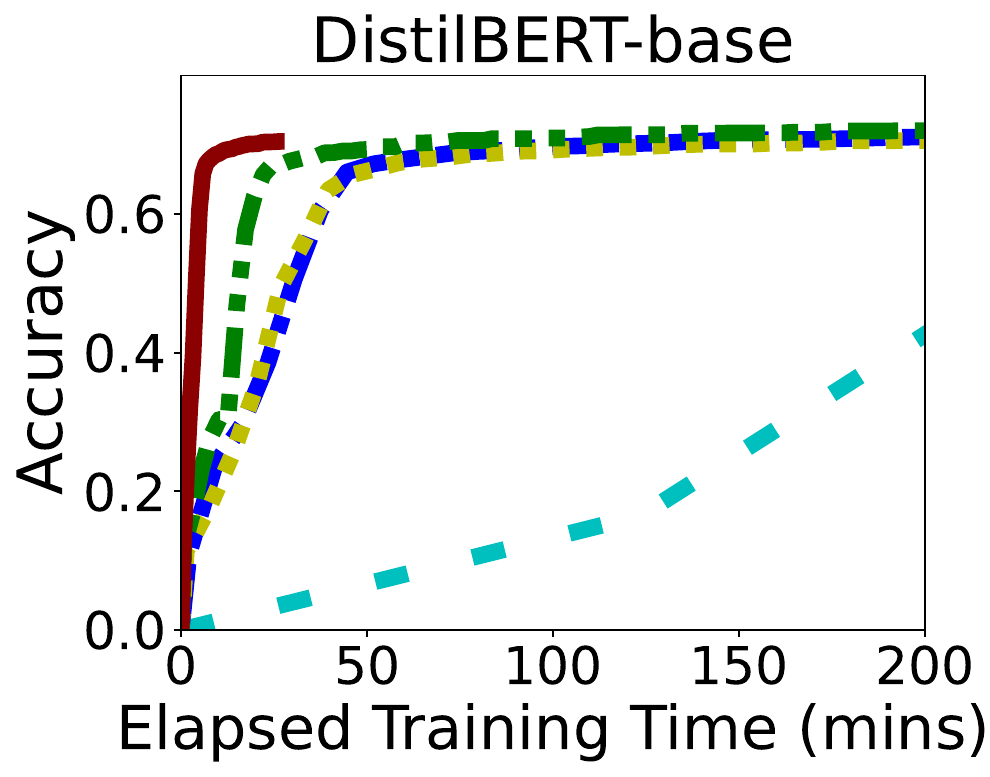}
        \end{minipage}
        ~
        \begin{minipage}[b]{0.24\textwidth}
            \includegraphics[width=1\textwidth]{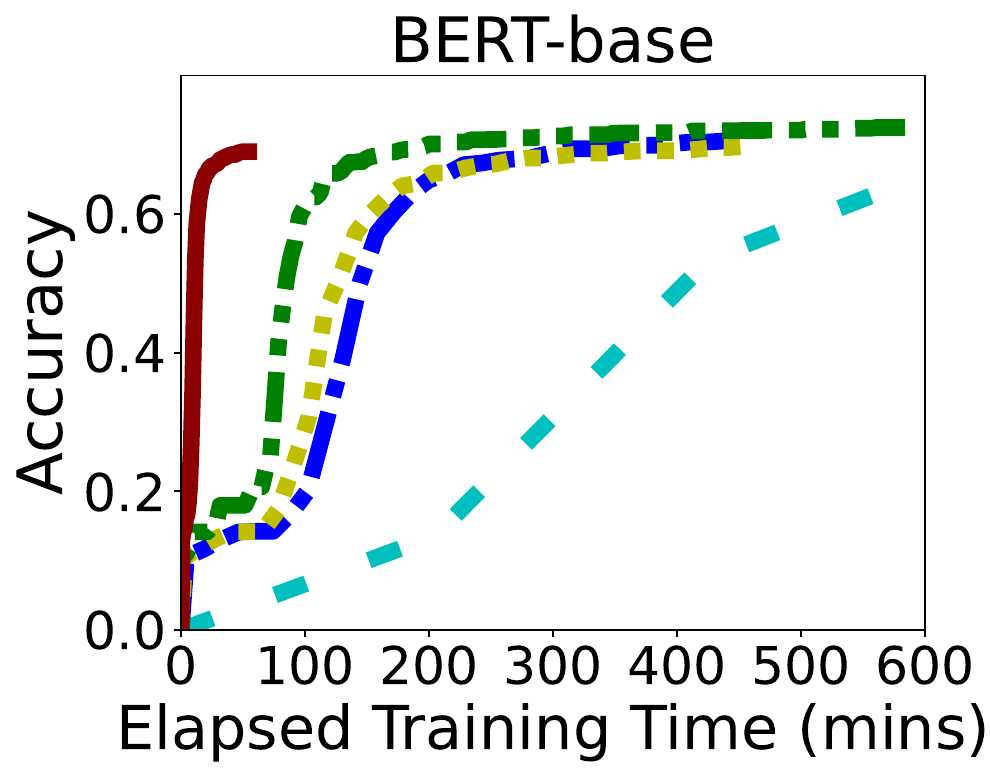}
        \end{minipage}
        ~
        \begin{minipage}[b]{0.24\textwidth}
            \includegraphics[width=1\textwidth]{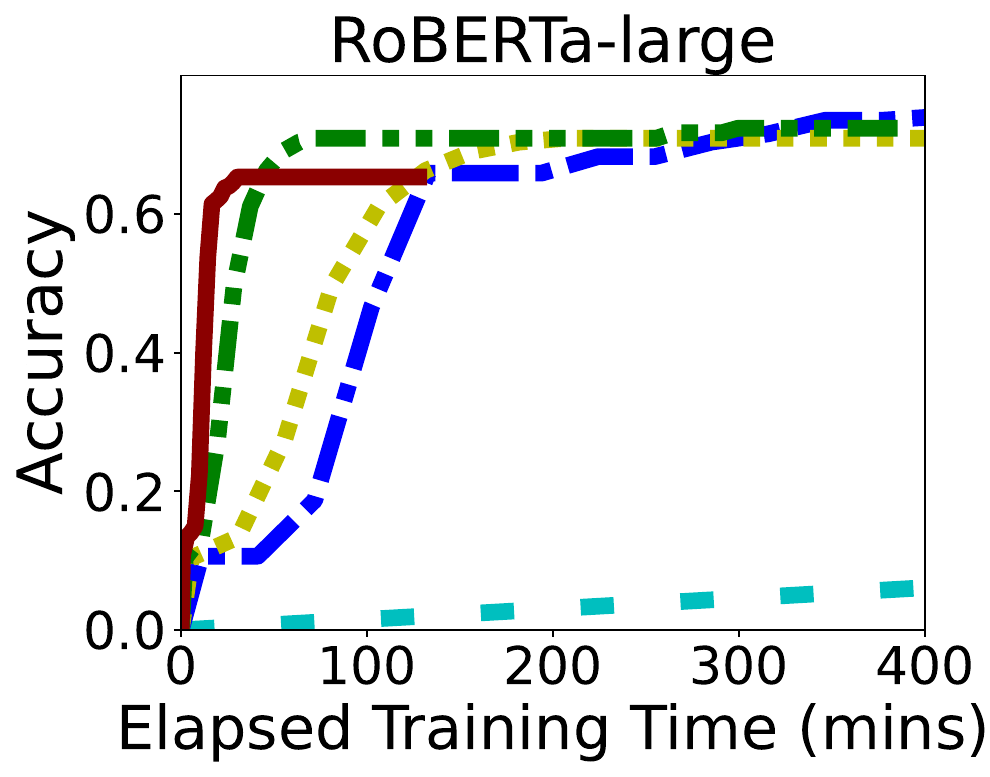}
        \end{minipage}\subcaption{\texttt{YAHOO}}
    \end{minipage}
    
    \begin{minipage}[b]{1\textwidth}
        \begin{minipage}[b]{0.24\textwidth}
            \includegraphics[width=1\textwidth]{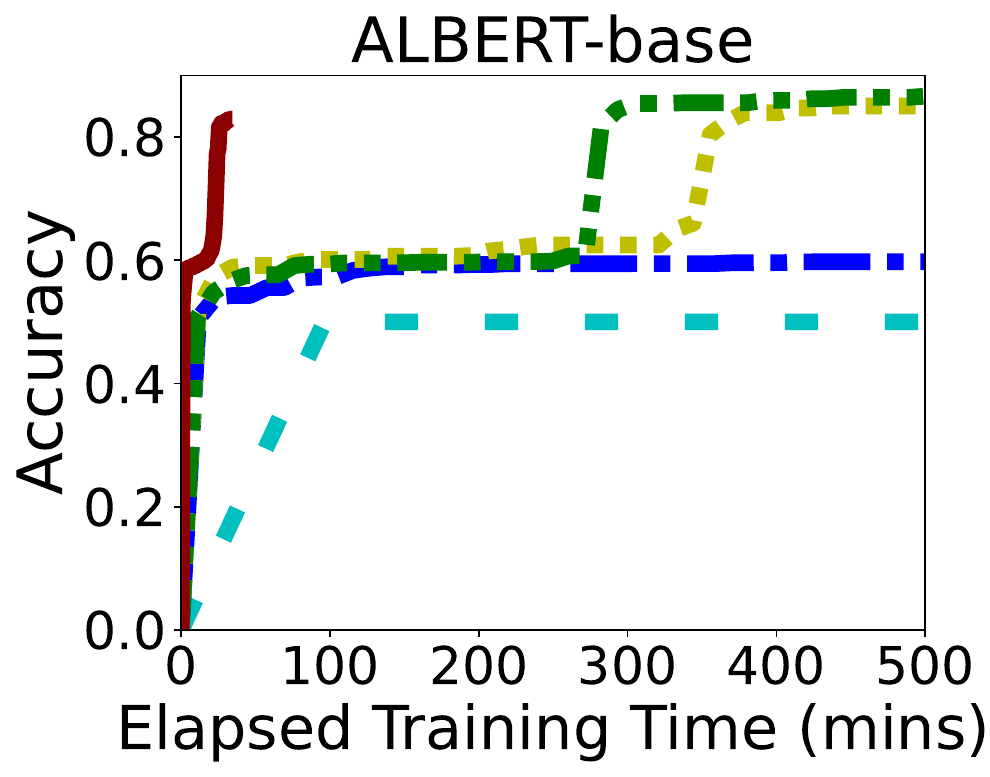}
        \end{minipage}
        ~
        \begin{minipage}[b]{0.24\textwidth}
            \includegraphics[width=1\textwidth]{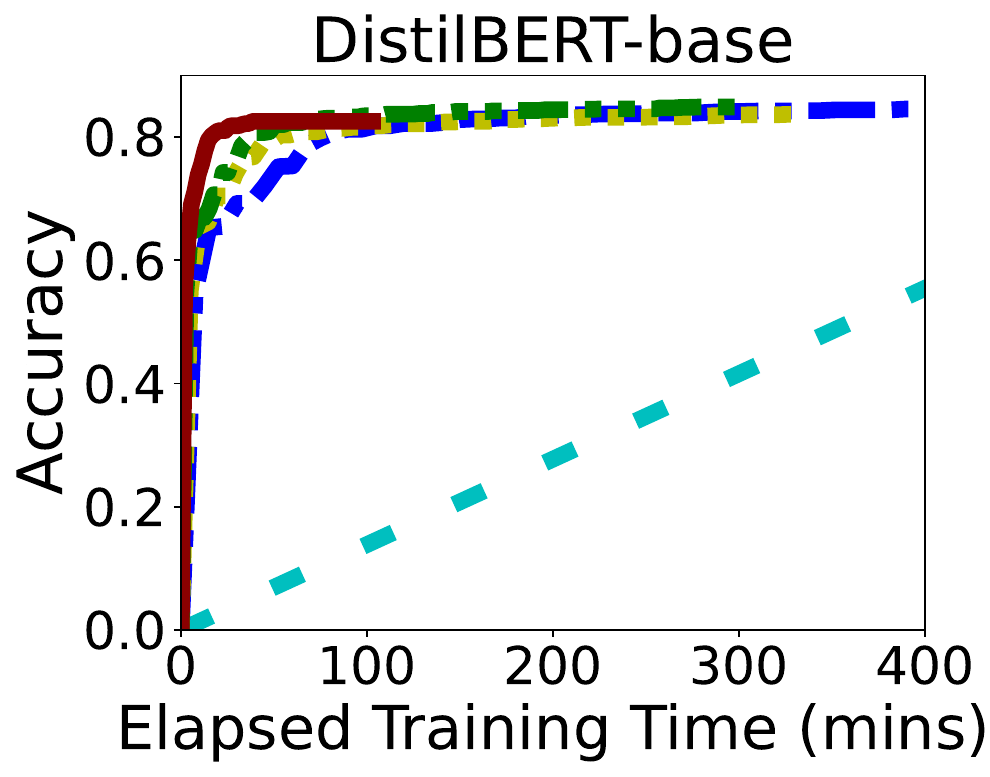}
        \end{minipage}
        ~
        \begin{minipage}[b]{0.24\textwidth}
            \includegraphics[width=1\textwidth]{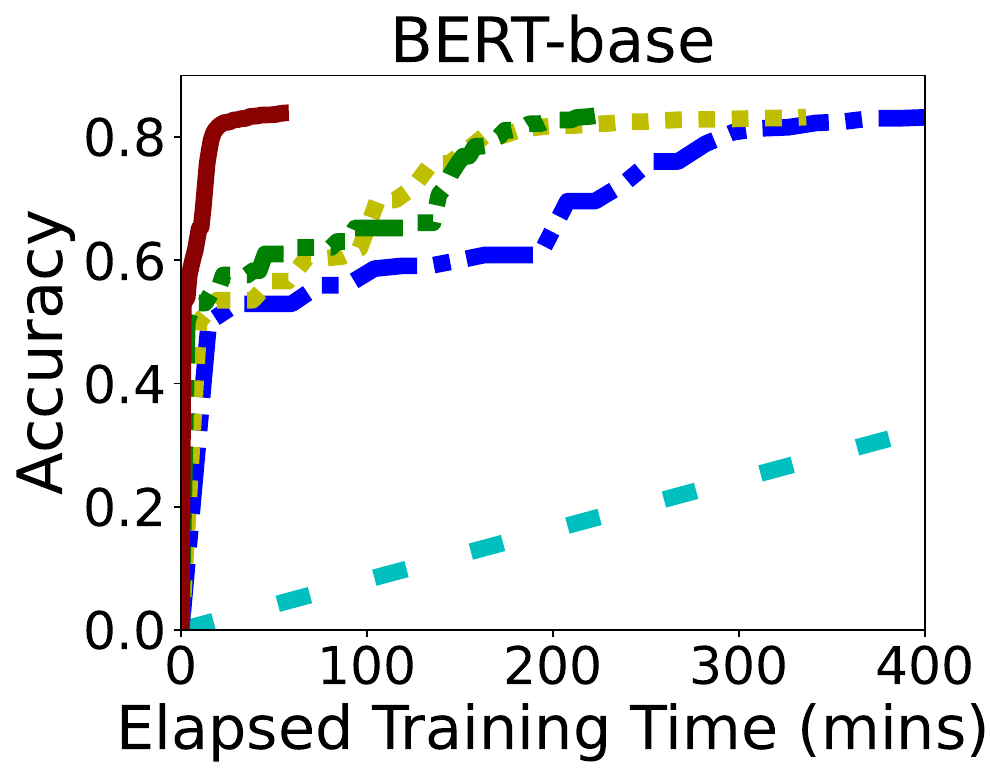}
        \end{minipage}
        ~
        \begin{minipage}[b]{0.24\textwidth}
            \includegraphics[width=1\textwidth]{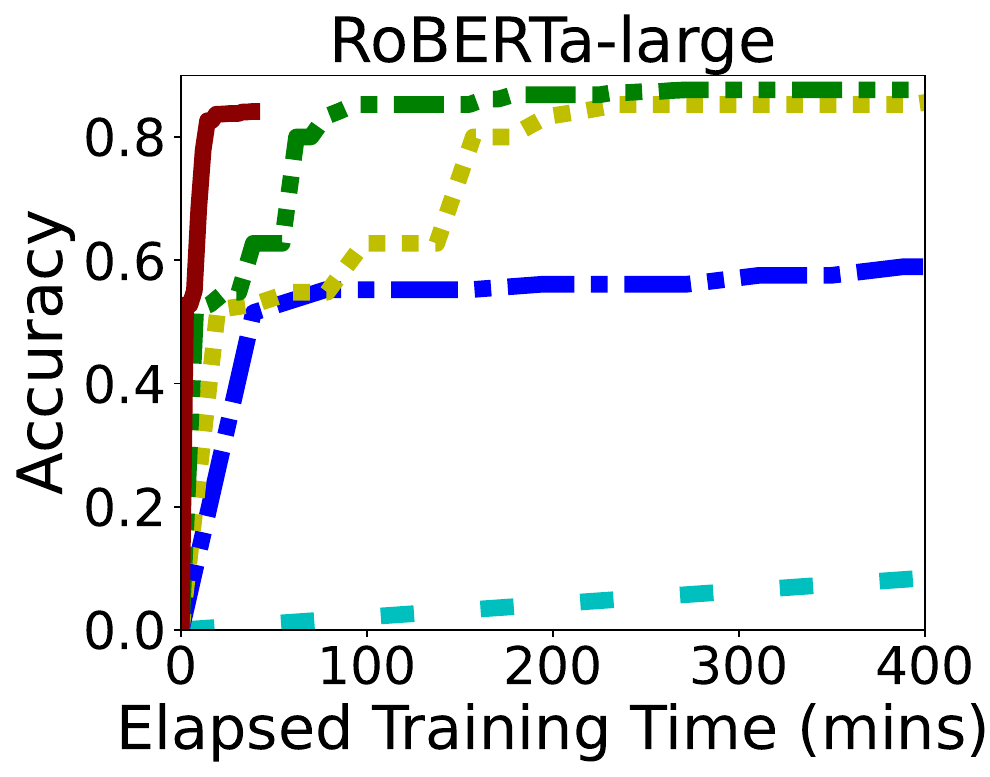}
        \end{minipage}\subcaption{\texttt{YELP-P}}
    \end{minipage}
    \vspace*{-20pt}

    \caption{Overall Performance of \sys and baselines. Processor: NPU for \sys and CPU for others.
    %We include 1000 clients per round for datasets \texttt{AGNEWS}, \texttt{YELP-P} and model \texttt{RoBERTa-large} for our \sys.
    	%LR: 0.1 for \texttt{AGNEWS} and 0.01 for others. 
    % \cdq{Q: What is the latency number used here?. really tested}
    }

    % acc_threhold = {
    % 'albert-agnews': 0.875,
    % 'distilbert-agnews':0.875,
    % 'bert-agnews':0.875,
    % 'roberta-large-agnews':0.875,
    % 'albert-yahoo':0.65,
    % 'distilbert-yahoo':0.7,
    % 'bert-yahoo':0.69,
    % 'roberta-large-yahoo':0.65,
    % 'albert-yelp-p':0.82,
    % 'distilbert-yelp-p':0.82,
    % 'bert-yelp-p':0.83,
    % 'roberta-large-yelp-p':0.82
    % }
    
    \label{fig:eval-overall}
\end{figure*}

\begin{table*}[t]
	\scriptsize
    \includegraphics[width=17.5cm]{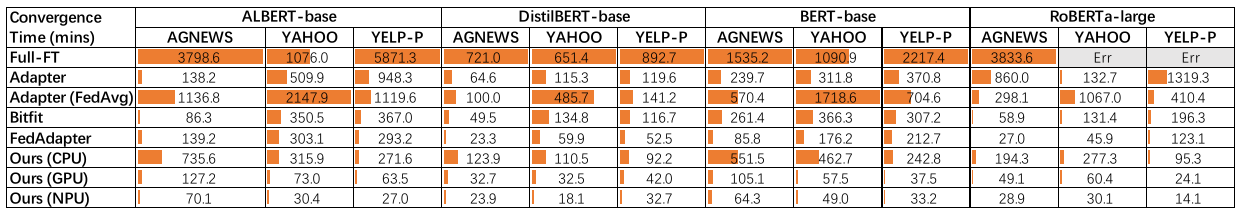}
\caption{Performance summary of Figure~\ref{fig:eval-overall} and its extension to different processors. Device: Google Pixel 7P.
% \cdq{to which accuracy?}
Err: failed to reach target accuracy within 100 hrs.
}
\label{tab:eval-e2e}
\vspace*{-10pt}
\end{table*}

\begin{comment}
We mainly evaluate the end-to-end performance of \sys to answer the following questions:
\textbf{Q1:}
How much performance improvement could \sys bring on powerful NLP processors?
\textbf{Q2:}
Could \sys still bring benefits when applied to common processors (i.e., CPU and GPU)?
\textbf{Q3:}
What is the scalability of \sys to different devices?
\textbf{Q4:}
How much benefit could \sys bring when involving different clients numbers?
\textbf{Q5:}
Is \sys resilient to non-iid data distribution?
\end{comment}

We first study the performance of \sys on the four BERT-like models that have less than 1B parameters and are extensively used in prior FedNLP research~\cite{cai2022augfedprompt,cai2022autofednlp,cai2023FeS,lin-etal-2022-fednlp}.

\noindent \textbf{\sys achieves significant improvements with mobile NPU.}
Table~\ref{tab:eval-e2e} summarizes the convergence time and Figure~\ref{fig:eval-overall} illustrates the convergence process under the default setting.
Specifically, we increase the involved client count to 1000 for some hard datasets and large model, so as to reach a higher performance.
% This is because, as previously shown in $\S$\ref{sec:back:exps}, 10 clients already saturate the convergence performance of backpropgation-based FL methods; while \sys is able to scale its convergence speed with more clients.
% Notably, not every device is training ready due to the runtime and battery status~\cite{gboard-fl}; yet 1,000 clients are commonly available in a popular mobile app.
% We will also compare the performance of \sys and baselines under different training-ready client numbers.
To reach the target accuracy, \sys outperforms \texttt{Full-FT} by an significant factor of 132.7$\times$.
Compared with parameter-efficient fine-tuning baseline and the state-of-the-art federated fine-tuning system FedAdapter~\cite{cai2022autofednlp}, \sys can still beat them non-trivially (9.6$\times$ on average).
The significant boost in performance partly stems from \sys's capacity to harness the powerful NPU, accelerating the training process.
Unique to \sys is its dependency solely on the forward pass, in contrast to other baselines that require the backward pass — a function not yet compatible with NPUs, as elaborated in Section~\ref{sec:back:exps}.

Intriguingly, our research reveals that, contrary to prior FL scenarios~\cite{li2021hermes,lipyramidfl,cai2022autofednlp}, FedLLM with PEFT methods favors FedSGD towards faster convergence compared to FedAVG.
As detailed in Table~\ref{tab:eval-e2e}, FedSGD using adapter leads to up to 3.3$\times$ faster convergence than FedAVG on the four tested models.
This can be rationalized by understanding the inherent design of FedAVG. Primarily devised to address the communication overhead in federated learning, the benefits of FedAVG become muted, as PEFT methods have already significantly mitigated these overheads.

\begin{figure}[t]
	\centering
    \includegraphics[width=0.45\textwidth]{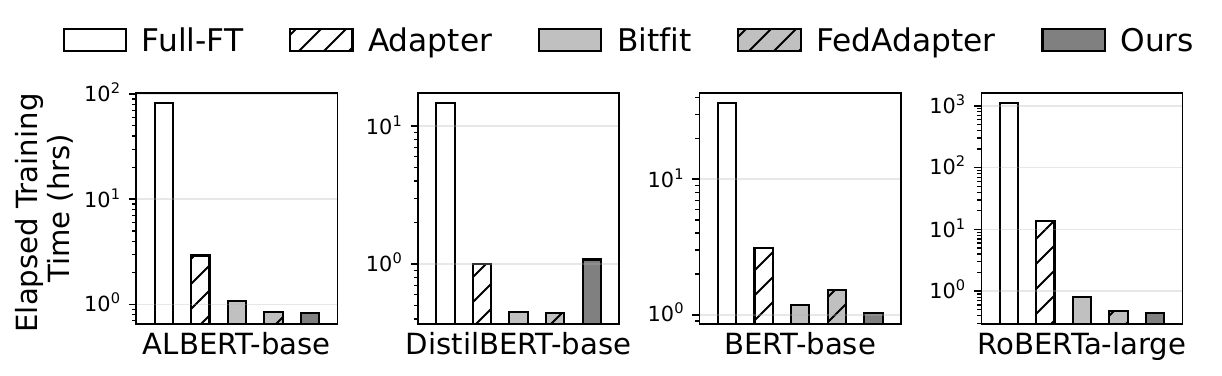}
    \caption{Convergence on heterogeneous hardware environment Jetson TX2 \& Pixel 7P for YELP-P.}
	\label{fig:eval-hardware}
	\vspace{-10pt}
\end{figure}

\noindent \textbf{\sys is versatile across different processors and hardware boards.}
Though designed primarily for NPUs, \sys showcases commendable performance across multiple processors within the Google Pixel.
On the GPU, it is up to 92.4$\times$ faster than the \texttt{Full-FT}, and it consistently outperforms several strong baselines, achieving speedup of 5.6$\times$ on average.
Remarkably, only \sys can exploit both the GPU and NPU in the Google Pixel, while other baselines are confined to the CPU, as illustrated in Table~\ref{tab:motivations-latency}.
Table~\ref{tab:eval-e2e} further showcases that, even on the Google Pixel's CPU, \sys can still operate up to 21.6$\times$ faster than \texttt{Full-FT} 
% \cdq{(DQ: clarify it is on average or up to?)}.
When compared to more advanced baselines, \sys outperforms 41.6\% of them and matches the remainder.
These results suggest that the design of \sys is not solely dependent on NPUs; it's equally effective on other processors.

%\noindent \textbf{\sys is scalable to different devices.}
Consistently, \sys surpasses other baselines regardless of the device or its underlying processor.
As evidenced in Figure~\ref{fig:eval-hardware}, when tested on 
the heterogeneous hardware environment, \sys maintains its lead. 
It stuck to the straggler latency.
Its convergence time showcases significant improvements, ranging from 12.2 to 3143.7$\times$ faster than \texttt{Full-FT}.
Furthermore, it beats strong baselines in most of the cases, delivering speeds up to 39.1$\times$ faster.
The advantage stems from the reality that a forward pass is inherently 3-5$\times$ quicker than a backward pass~\cite{baydin2022gradients, blayo2014advanced}.
The numerical gap between forward and backward pass are highly program dependent, which will be exaggerated in efficient inference engines~\cite{liu2023efficientvit, liang2023dynamic, vatsavai2023optical}.
Apart from that, our global cherry-picked forward gradients, refined through discriminative filtering, help to guide the training process to reach the target accuracy quickly.

\begin{figure}[t]
	\centering
        \includegraphics[width=0.25\textwidth]{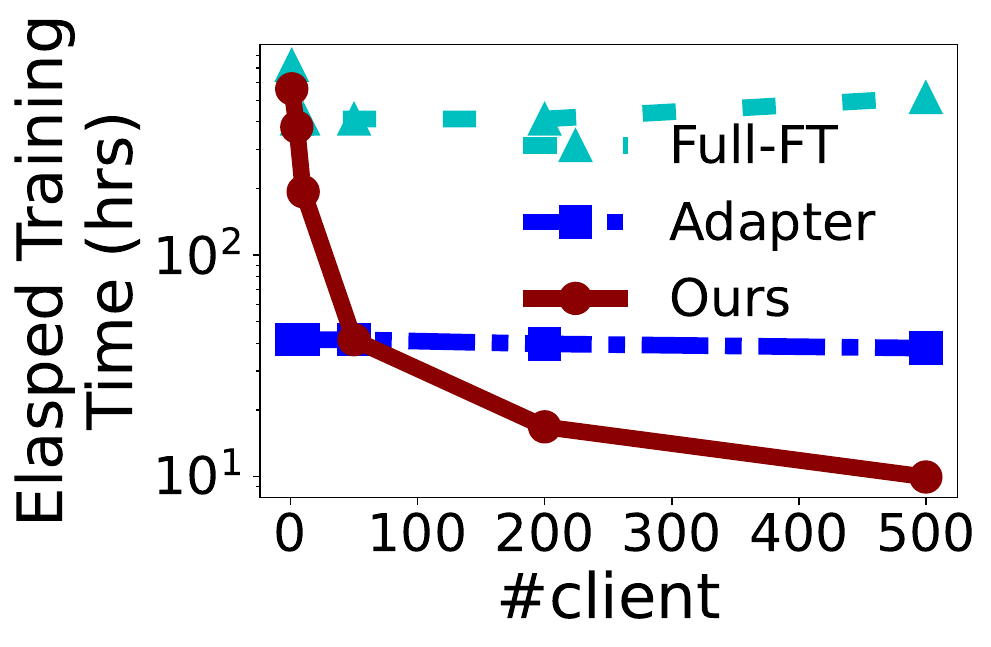}\vspace*{-3.5pt}
                        \caption{\sys performance with different \#clients. 
                        % Dataset: AGNews; Model: DistilBERT-base; Target Accuracy: 0.86.
                        % \cdq{Remove fedavg results to avoid confusion. Include baselines results.
                        }
                        \label{fig:eval-model-bert-large}
\end{figure}

\begin{figure}[t]
        \centering
        \includegraphics[width=0.3\textwidth]{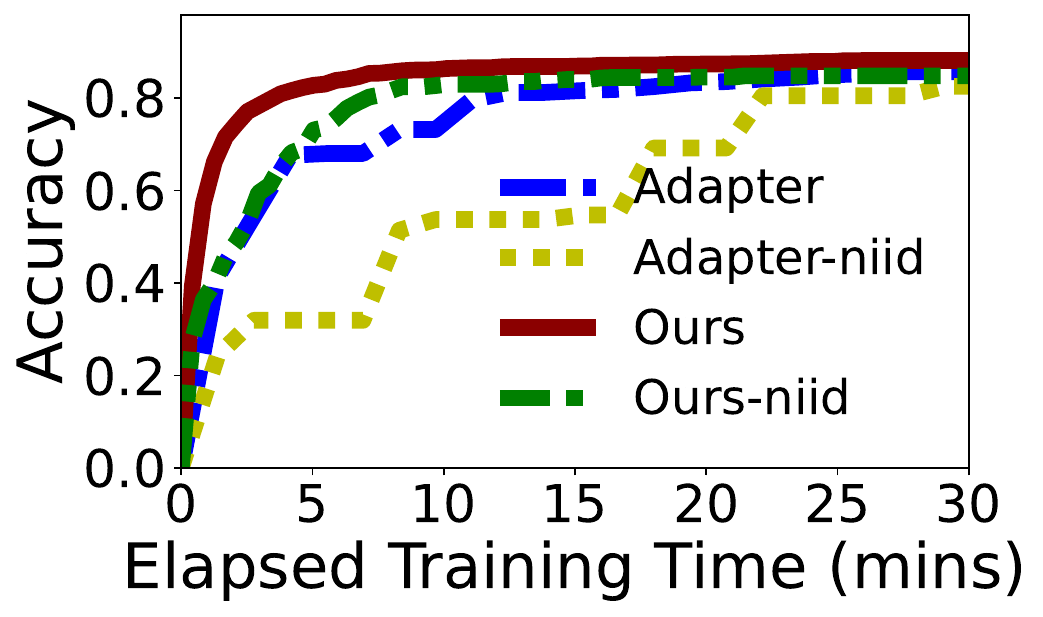}
        \caption{Non-iid performance for \texttt{AGNEWS} on \texttt{DistilBERT}.}
        \label{fig:eval-niid}
\end{figure}

% \begin{figure}[t]
% 	\centering
%     \includegraphics[width=0.23\textwidth]{figs/eval-e2e-clients.pdf}
%     % 这里fedavg和fedsgd用的其实是bert模型，distilbert等后续重跑
%     \caption{End-to-end performance under different involved clients. 
%     % \cdq{Remove fedavg results to avoid confusion. Include baselines results.}
%     }
% 	\label{fig:eval-model-bert-large}
% \end{figure}

% \begin{figure}[t]
% 	\centering
%     \includegraphics[width=0.23\textwidth]{figs/eval-llama-agnews.pdf}
%     \caption{the performance of \sys on LLaMA. 
%     % \cdq{Change the axis to real runtime (quantized). 
%     % The meta-data could be taken from MobileFM (in new Jetson board).
%     % The baseline is backpropagation training in NVIDIA A100.
%     % Forward-FL BS = 8.
%     % Client = 100.
%     % LR = 3e-4.}
%     }

% 	\label{fig:eval-llama-agnews}
% \end{figure}
\noindent \textbf{\sys exhibits enhanced scalability in convergence speed as the number of available training devices increases.}
Figure~\ref{fig:eval-model-bert-large} demonstrates that \sys adeptly leverages a growing number of clients to bolster convergence performance.
With the client count increasing from 1 to 500, \sys's convergence duration notably reduces from 563.65 to 9.96 minutes.
On the other hand, vanilla FedLLM methods, which rely on backward passes, fail to showcase a similar trend in scalability.
For these approaches, the convergence duration plateaus once the client count exceeds 5.
This behavior echoes the findings in $\S$\ref{sec:back:exps}, underscoring that \sys outperforms backward pass-based FedLLM methods in scalability as more clients are added.

% \input{fig-eval-niid.tex}
% \usepackage{multirow}
% \usepackage{graphicx}
\begin{table}[]
\centering
\resizebox{0.85\columnwidth}{!}{%
\begin{tabular}{|l|ll|ll|}
\hline
\multirow{2}{*}{\begin{tabular}[c]{@{}l@{}}\textbf{Convergence} \\ \textbf{time (mins)}\end{tabular}} & \multicolumn{2}{l|}{\textbf{DistilBERT-base}}       & \multicolumn{2}{l|}{\textbf{BERT-base}}             \\ \cline{2-5} 
                                                                           & \multicolumn{1}{l|}{Uniform} & non-IID & \multicolumn{1}{l|}{Uniform} & non-IID \\ \hline
\texttt{Adapter}                                                                    & \multicolumn{1}{l|}{23.4}    & 37.8   & \multicolumn{1}{l|}{92.3}    & 158.2  \\ \hline
\multicolumn{1}{|l|}{{\color[HTML]{8B0000} Ours}}                                                                        & \multicolumn{1}{l|}{{\color[HTML]{8B0000} {9.1}}}  & \multicolumn{1}{l|}{{\color[HTML]{8B0000} {11.2}}} & \multicolumn{1}{l|}{{\color[HTML]{8B0000} {28.4}}} & \multicolumn{1}{l|}{{\color[HTML]{8B0000} {51.2}}}
\\ \hline
\end{tabular}%
}
\caption{Summary of FwdFL performance under non-iid data distribution. Target accuracy is set as 0.84.}
\label{tab:eval-niid}
\vspace*{-10pt}
\end{table}

\noindent \textbf{\sys demonstrates resilience to non-iid data distributions.}
We evaluate the performance of \sys under non-iid data distribution, as presented in Figure~\ref{fig:eval-niid}.
Notably, while the performance of \sys under non-iid slightly lags behind its IID counterpart — witnessing up to a 3.8\% accuracy drop in AGNEWS dataset, it still achieves parity convergence accuracy with strong baseline methods under non-iid circumstances, with up to 337.5\% faster.
The resilience of \sys to non-iid data distributions can be attributed to its ability to engage a vast number of clients, effectively harnessing their collective knowledge in a single round.

\subsection{Convergence Performance on Billion-Sized Model (LLaMA)}
\label{sec:eval-llama}
% 这个图是用的真实的gpu time，或许要想办法转成fl的时间或者直接横坐标换成step?

% \usepackage{multirow}
% \usepackage{graphicx}
\begin{table}[]
	\resizebox{0.98\columnwidth}{!}{%
		\begin{tabular}{|l|r|rrr|ccc|}
			\hline
			\multicolumn{1}{|c|}{} &
			\multicolumn{1}{c|}{} &
			\multicolumn{3}{c|}{\textbf{Centralized Training (A100)}} &
			\multicolumn{3}{c|}{\textbf{Federated Learning}} \\ \cline{3-8} 
			\multicolumn{1}{|c|}{\multirow{-2}{*}{\textbf{Methods}}} &
			\multicolumn{1}{c|}{\multirow{-2}{*}{\textbf{\begin{tabular}[c]{@{}c@{}}Mem.\\ (GB)\end{tabular}}}} &
			\multicolumn{1}{c|}{\textbf{Acc.}} &
			\multicolumn{1}{c|}{\textbf{Round}} &
			\multicolumn{1}{c|}{\textbf{Time}} &
			\multicolumn{1}{c|}{\textbf{Acc.}} &
			\multicolumn{1}{c|}{\textbf{Round}} &
			\textbf{Time} \\ \hline
			BP, FP16 &
			39.2 &
			\multicolumn{1}{r|}{89.7} &
			\multicolumn{1}{r|}{500} &
			0.1 hrs &
			\multicolumn{3}{c|}{} \\ \cline{1-5}
			BP, INT8 &
			32.4 &
			\multicolumn{1}{r|}{88.6} &
			\multicolumn{1}{r|}{500} &
			0.06 hrs&
			\multicolumn{3}{c|}{} \\ \cline{1-5}
			BP, INT4 &
			28.5 &
			\multicolumn{1}{r|}{87.8} &
			\multicolumn{1}{r|}{500} &
			0.04 hrs&
			\multicolumn{3}{c|}{} \\ \cline{1-5}
			Ours, FP16 &
			15.6 &
			\multicolumn{1}{r|}{87.0} &
			\multicolumn{1}{r|}{240} &
			1.5 hrs&
			\multicolumn{3}{c|}{} \\ \cline{1-5}
			Ours, INT8 &
			7.9&
			\multicolumn{1}{r|}{86.9} &
			\multicolumn{1}{r|}{260} &
			0.8 hrs&
			\multicolumn{3}{c|}{\multirow{-5}{*}{\begin{tabular}[c]{@{}c@{}}N/A due to memory\\ inefficiency on\\ Pixel 7 Pro (8GB)\end{tabular}}} \\ \hline
			{ Ours (CPU), INT4} &
			 &
			\multicolumn{1}{r|}{} &
			\multicolumn{1}{r|}{} &
			 &
			\multicolumn{1}{r|}{{ }} &
			\multicolumn{1}{r|}{{ }} &
			\multicolumn{1}{r|}{{ 0.19 hrs}} \\ 
			{  Ours (NPU$^*$), INT4} &
			{  \multirow{-2}{*}{\begin{tabular}[c]{@{}c@{}}4.0\end{tabular}} } &
			\multicolumn{1}{r|}{{  \multirow{-2}{*}{\begin{tabular}[c]{@{}c@{}}85.8\end{tabular}}}} &
			\multicolumn{1}{r|}{{  \multirow{-2}{*}{\begin{tabular}[c]{@{}c@{}}130\end{tabular}}}} &
			{  \multirow{-2}{*}{\begin{tabular}[c]{@{}c@{}}0.25 hrs\end{tabular}}} &
			\multicolumn{1}{r|}{{  \multirow{-2}{*}{\begin{tabular}[c]{@{}c@{}}85.8\end{tabular}}}} &
			\multicolumn{1}{r|}{{  \multirow{-2}{*}{\begin{tabular}[c]{@{}c@{}}130\end{tabular}}}} &
			\multicolumn{1}{r|}{{  0.07 hrs}} \\ \hline
		\end{tabular}%
	}
	\caption{
	\sys combined with INT4-based quantization is the only feasible approach in federated learning of LLaMA-7B. Dataset: AGNEWS; Acc.: accuracy (\%). Centralized training and federated learning are conducted on NVIDIA A100 and Pixel 7 Pro, respectively. *: LLaMA currently is not supported by mobile NPU, therefore we emulate its speed based on the speedup of other BERT-like models.}
\label{tab:eval-llama}
\vspace*{-10pt}
\end{table}

\textbf{Discriminative task.}
We further assess \sys using a billion-sized model, LLaMA-7B, on the AGNEWS dataset and the Pixel 7 Pro.
To accommodate the model on these devices, we employed the quantization technique GPTQ~\cite{frantar2022gptq}, which is prevalently utilized with LLMs to minimize parameter redundancy.
During the fine-tuning process, we adopted low-precision data formats like INT8/INT4 for the native LLaMa weights, while retaining FP32 for the trainable LoRa weights.
Given that backpropagation-based training approaches are not feasible for such sizable models on mobile platforms, we conducted experiments using both centralized training (leveraging 1x NVIDIA A100) and federated learning (on smartphones).

With results shown in Table~\ref{tab:eval-llama}, we make following key observations.
(1) For the first time, \sys enables federated learning of billion-sized LLMs like LLaMA on COTS mobile devices.
Combined with INT4 quantization, it takes only 0.19 hours for \sys to achieve the target accuracy running on mobile CPU, which is comparable to a centralized training with one NVIDIA A100 GPU.
This is mainly due to its ability to scale out its speed on a thousand devices simtaneously.
If on-device training can be accelerated by NPU, \sys is even much faster than the centralized training.
%Note that in the centralized training, our approach converges much slower than backpropagation-based approach since our implementation cannot batch the training of 
%  with a comparable speed, which mainly attributes to the crowded clients engaged.
(2) \sys exhibits a harmonious orchestration with quantization strategies.
Quantizing LLaMA weights to INT8 and INT4 effectively reduces the training-time memory footprint from 15.6GBs to 7.9GBs and 4.0GBs.
Yet, the accuracy only degrades by less than 1.2\%.
This shows the great compatibility of \sys with existing model compression algorithms.

\begin{figure}[t]
	\centering
    \includegraphics[width=0.45\textwidth]{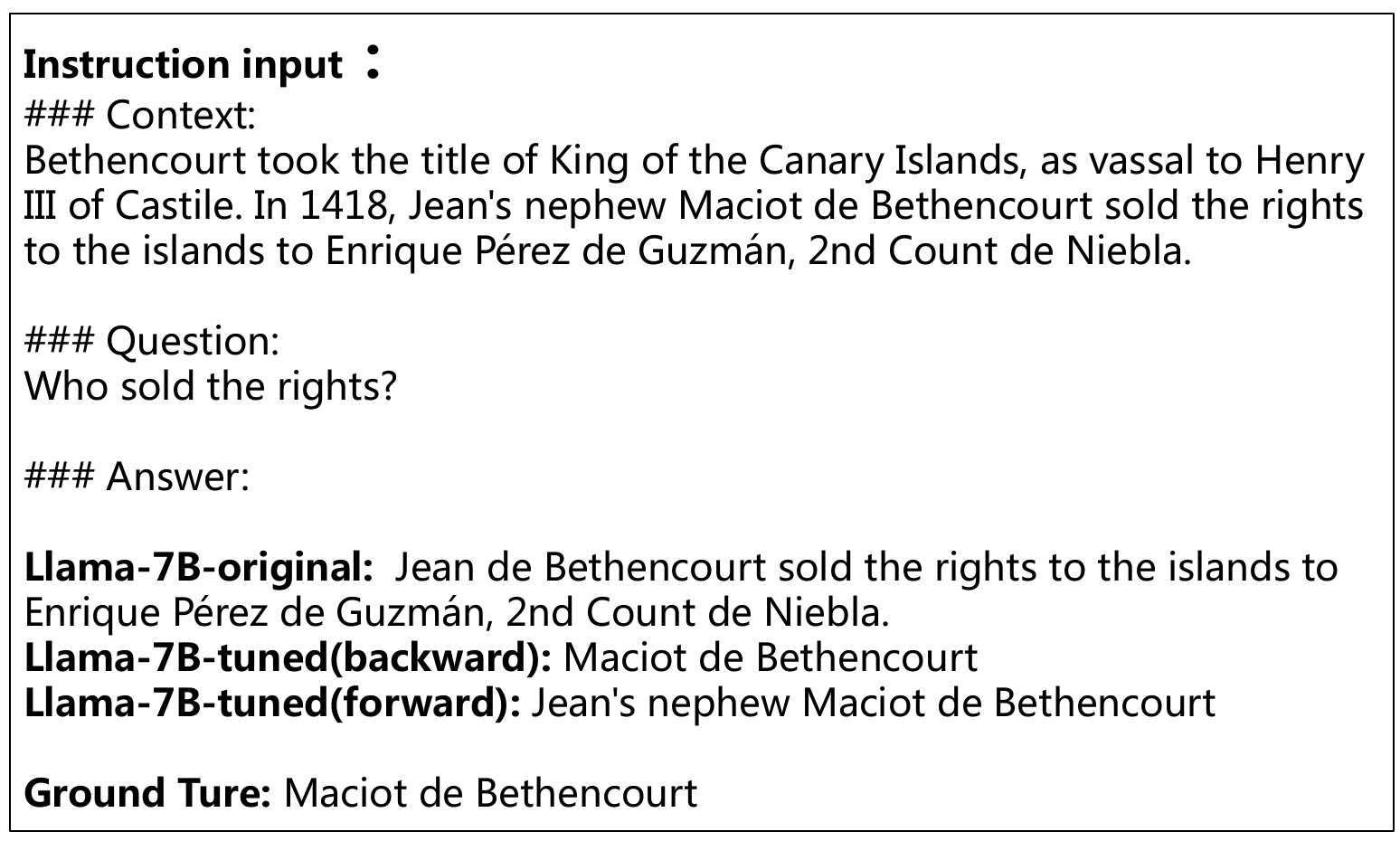}
    \caption{A case showing how \sys guides the LLaMA to follow human instructions.}
	\label{fig:eval-llama-squad}
\end{figure}
\textbf{Generative (instruction-following) task.}
We also evaluate \sys on the generative task \texttt{SQUAD}.
As shown in the Figure~\ref{fig:eval-llama-squad}, \sys could generate similar results as the centralized training.
Statistically, \sys could reach 83\% f1-score after 312 rounds instruct tuning, which is 55.2\% higher than zero-shot performance.
Even compared with centralized BP-based training, \sys is only 
1.9\% lower, which do not affect the generative quality.
Note that \sys conducts experiments on the federated setting with only forward pass, which is private and fast as we depicted in Table~\ref{tab:eval-llama}.

\subsection{System Cost}
\label{sec:eval-cost}

\begin{figure}[t]
	\centering
	\begin{minipage}[b]{0.48\textwidth}
		\includegraphics[width=1\textwidth]{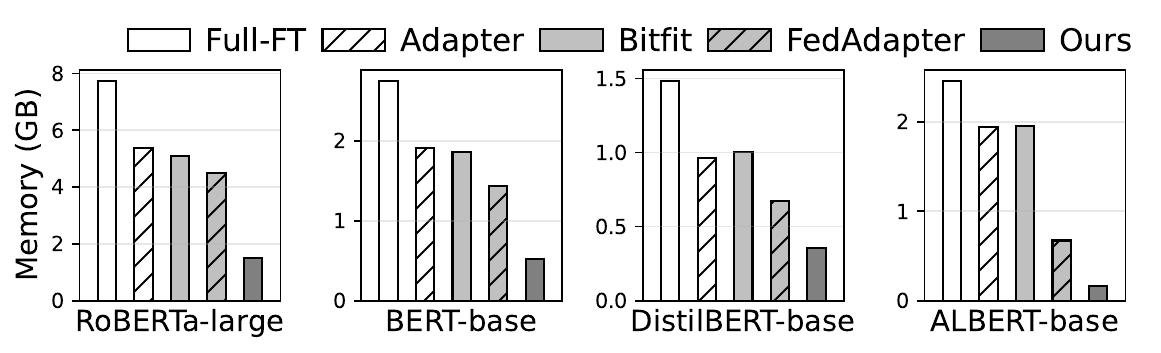}
		\subcaption{Peak memory footprint}
		\label{fig:eval-cost-memory}
	\end{minipage}
    \begin{minipage}[b]{0.23\textwidth}
        \includegraphics[width=2\textwidth]{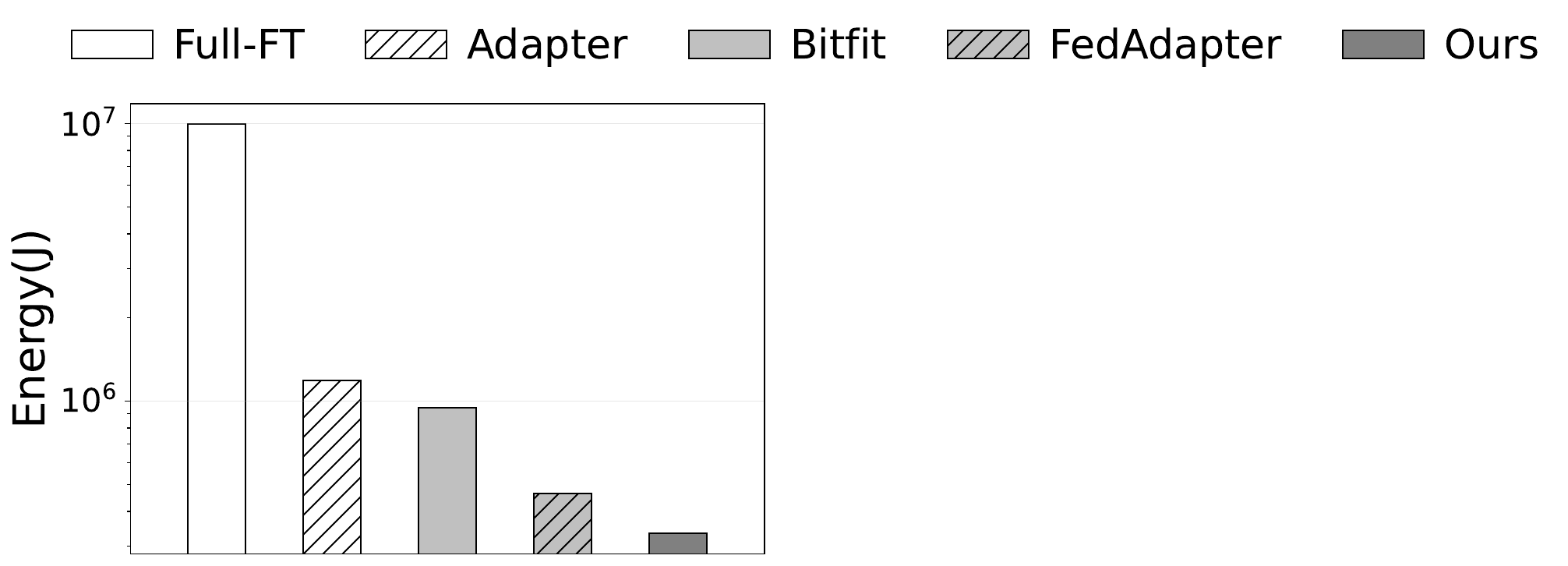}
        \subcaption{Total energy cost}
        \label{fig:eval-cost-energy-all}
    \end{minipage}~
    \begin{minipage}[b]{0.231\textwidth}
        \includegraphics[width=1\textwidth]{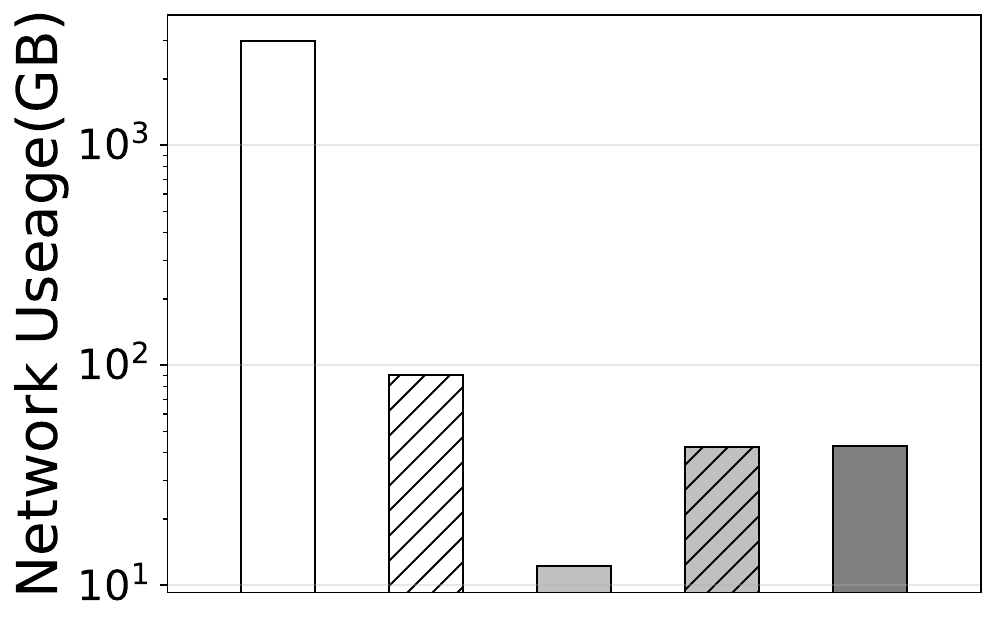}
        \subcaption{Total network cost}
        \label{fig:eval-cost-network-all}
    \end{minipage}
    \vspace*{-10pt}
    \caption{Resource cost of \sys and baselines.} % Model: \mwx{ALBERT}; Dataset: \mwx{YELP-P} for energy and \mwx{RobBERTa-large}; Dataset: \mwx{YELP-P} for momery.}
    \label{fig:eval-cost}
    % \vspace*{-20pt}
    
\end{figure}

We analyze the resource cost during FedLLM, including the peak memory footprint, total energy and network (both uplink and downlink) expenditure on all participant devices.
The experiments are performed with RoBERTa-large and YELP-P on Pixel 7 Pro.
The results are depicted in Figure~\ref{fig:eval-cost}.

% \begin{figure}[t]
% 	\centering
%     \includegraphics[width=0.25\textwidth]{figs/eval-per-client-system-cost-bert-agnews.pdf}
%     \caption{Cost 1}
% 	\label{fig:eval-cost-energy-client}
% \end{figure}

% Though we use more clients, we
% save comm because PEFT downloading and scales uploading and alleviate per client energy consumption, improve average QoS.
% Most importantly, we significantly reduce the memory usage, which is the main bottleneck of running LLMs on mobile devices~\cite{wang2022melon}.

\paragraph{Memory footprint}
As shown in Figure~\ref{fig:eval-cost-memory}, \sys achieves up to a 93\% reduction in memory usage compared to FT and a 91.3\% decrease relative to the robust PEFT benchmarks.
Notably, only 169MB of memory is required for a single \sys round, rendering it highly suitable for mobile devices.
This efficiency stems from \sys's design, which mandates only the execution of the forward pass, thus requiring storage primarily for the trainable model parameters and the perturbation weights; both are parameter-efficient as detailed in $\S$\ref{sec:design-overview}.
Moreover, due to \sys's sole reliance on inference functions, cutting-edge inference-centric memory optimization techniques~\cite{liu2023efficientvit, liang2023dynamic, vatsavai2023optical} can be effortlessly incorporated to further reduce memory consumption.

\paragraph{Energy and network}
%Figure~\ref{fig:eval-cost-energy-all} on AGNEWS demonstrates that \sys reduces network consumption by up to \cdq{99.17}\% compared to FT, yet exhibits network consumption comparable to the robust PEFT baselines.
Compared to \texttt{Full-FT}, \sys is able to save 96.7\% energy consumption and 98.6\% network cost across devices.
\sys is still network-efficient compared to competitive baselines with PEFT enhancements, e.g.,  2.1$\times$ less network traffic compared to Adapter approach.%后面1.5和2.1是干啥的来着。。adapter和bitfit取平均值吗，要加上fedadapter吗; 就是和adapter比
The only expectation is BitFit because it only transmits bias of each parameter in the network, i.e., only 0.1\% of the total parameters.
% \cdq{We note that we could only transmit the scale loss according to FedKSeed.}
In the aspects of energy consumption, \sys consumes 2.6$\times$ on average less energy than all the PEFT baselines.
The main reason behind that is \sys only need to compute forward pass, which is  3-5$\times$ quicker than a backward pass~\cite{baydin2022gradients, blayo2014advanced} and could be accelerated by NPU processors.

% that it only involves a forward pass, which is 3-5$\times$ quicker than a backward pass~\cite{baydin2022gradients, blayo2014advanced}.
% It is mainly ascribed to the involvement of 100$\times$ more devices per round, i.e., a price paid for scalability.
% Though, we deem such a price paid is bearable since FL participant devices are typically in a good condition, i.e., with battery being charged and non-metered nework connectivity like WiFi~\cite{gboard-fl}.

%\paragraph{Network Traffic}
% \cdq{Shall multiply by 2 because we upload gradients now.}
%As depicted in Figure~\ref{fig:eval-cost-network-all}, \sys achieves more than 99\% reduction in network consumption when compared to full fine-tuning, and its consumption parallels that of the established PEFT benchmarks.
%On public cloud platforms, network costs are typically invoiced based on data transmission volume, such as a rate of \$0.01/GB on AWS~\cite{awsbill2022}.
%By reducing network traffic, both the economic burden on clients and the financial outlays for FL developers are mitigated.

% \mwx{Where is memory cost?}

\subsection{Significance of key designs}
\label{sec:eval-ablation}

\begin{figure}[t]
	\centering
    \includegraphics[width=0.48\textwidth]{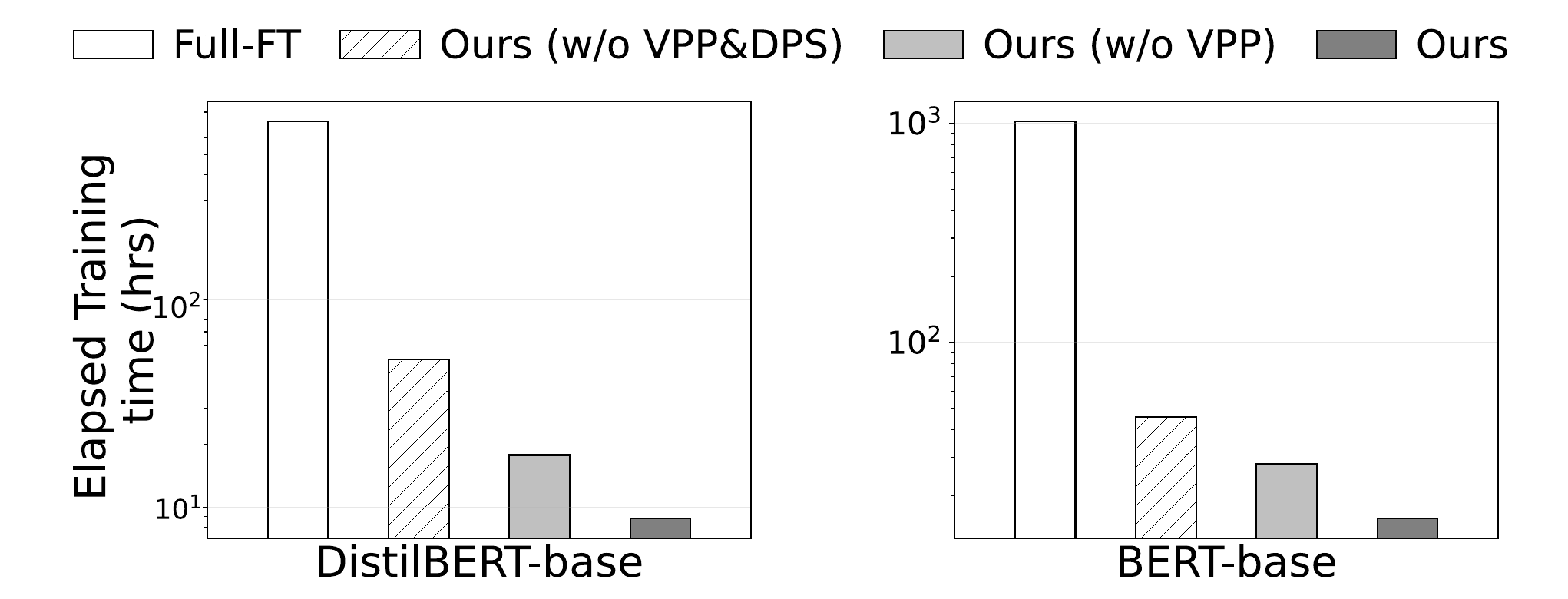}
    \vspace*{-20pt}
    \caption{Model convergence delay with or without \sys's key designs, showing their significance.
    VPP means \textbf{V}ariance-controlled \textbf{P}erturbation \textbf{P}acing, and DPS means \textbf{D}iscriminative \textbf{P}erturbation \textbf{S}ampling.
    % \cdq{Client: 100. Model: DistilBERT.}
    % \mwx{need 2 subfigures for this.}
    % 因为用的v是之前的10倍，所以vanilla forward-fl的效果一般
    }
    % \vspace*{-20pt}
	\label{fig:eval-ablation}
    \vspace*{-10pt}
\end{figure}

\paragraph{Dissecting system benefits}
Figure~\ref{fig:eval-ablation} reveals that each design element plays a pivotal role in enhancing the performance of \sys.
(1). 
Vanilla Forward-FL (\sys without variance-controlled perturbation and discriminative perturbation sampling) dramatically reduces the convergence time, ranging from 14.13$\times$ to 22.45$\times$. 
This improvement is ascribed to \sys's capability to engage more clients in local PEFT training, thereby magnifying their contributions during the training phase.
(2). 
Discriminative perturbation sampling can augment the convergence rate by 1.63$\times$ to 2.88$\times$. This enhancement is realized by harnessing the full potential of each perturbation, enabling them to contribute more significantly to the gradient updates.
(3). 
Variance-controlled perturbation pacing further bolsters convergence performance, with improvements ranging from 1.78$\times$ to 2.02$\times$. This can be attributed to two primary factors: the early elimination of ineffective perturbations and the precise identification of optimal perturbations in subsequent phases.

\begin{figure}[t]
	\centering
    \begin{minipage}[b]{0.23\textwidth}
        \includegraphics[width=1\textwidth]{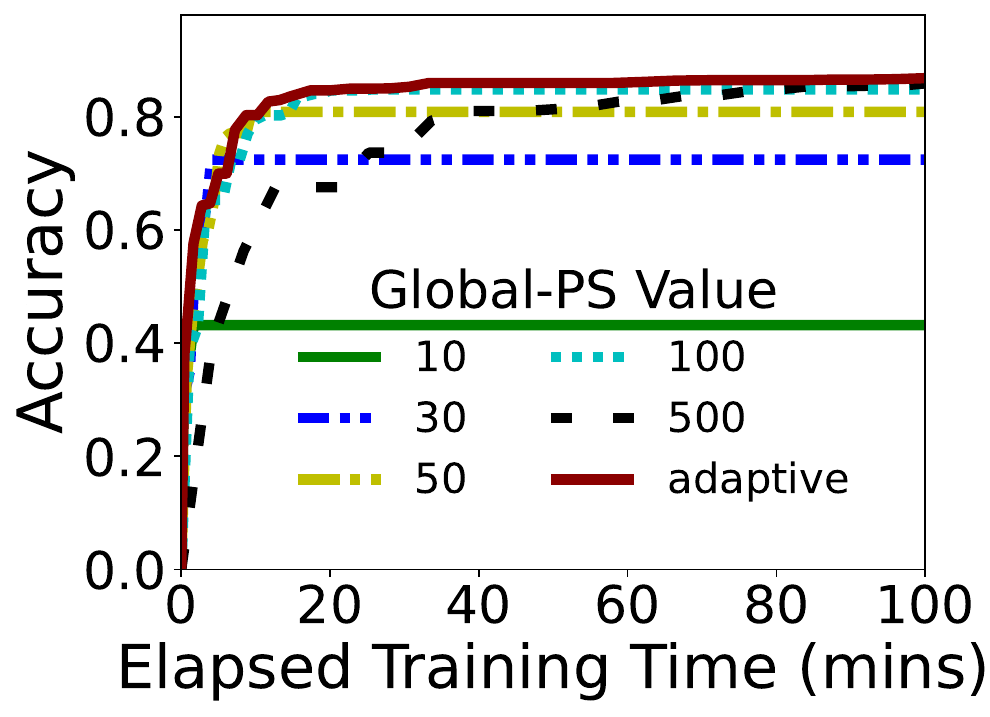}
        \vspace*{-20pt}
        \caption{Adaptive vs Fixed. Model:DistilBERT. Dataset:AGNews}
        \label{fig:eval-ablation-adaptive}
    \end{minipage}~\hspace*{3pt}
    \begin{minipage}[b]{0.23\textwidth}
        \includegraphics[width=1\textwidth]{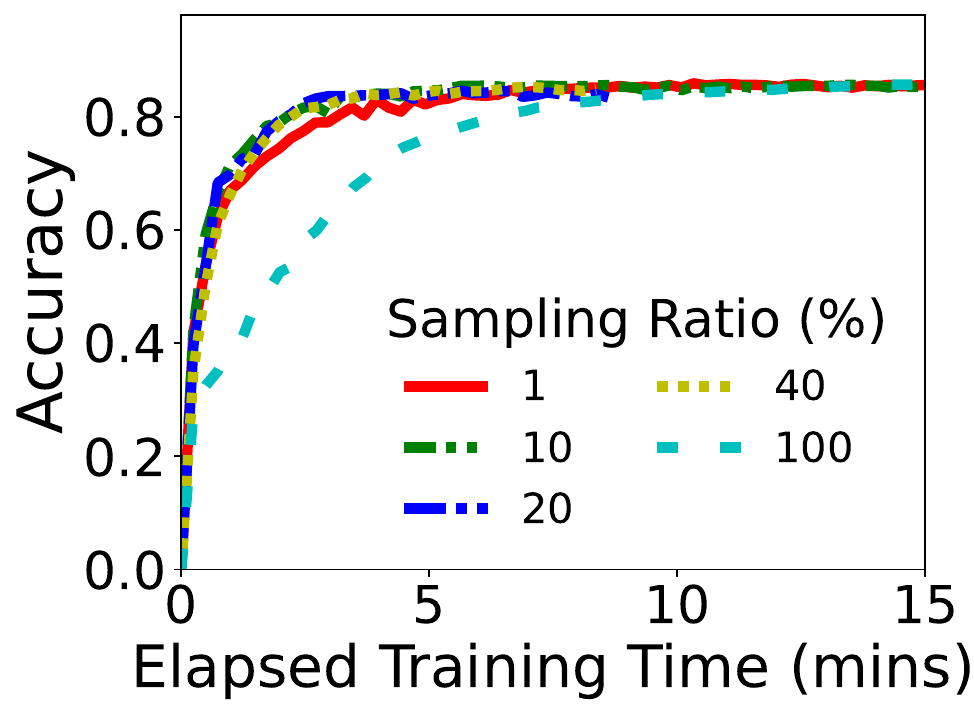}
        \vspace*{-20pt}
        \caption{Different sampling ratio. Model:BERT. Dataset:AGNews}
        \label{fig:eval-ablation-sampling}
    \end{minipage}
    % \caption{Different sampling ratio.}
	% \vspace*{-20pt}
\end{figure}

% \begin{figure}[t]
% 	\centering
%     \begin{minipage}[b]{0.24\textwidth}
%         \includegraphics[width=1\textwidth]{figs/eval-ablation-sampling-bert-agnews.pdf}
%         \subcaption{Model 1}
%     \end{minipage}~
%     \begin{minipage}[b]{0.23\textwidth}
%         \includegraphics[width=1\textwidth]{figs/eval-ablation-sampling-distilbert-agnews.pdf}
%         \subcaption{Model 2}
%     \end{minipage}
%     \caption{Different sampling ratio.}
% 	\label{fig:eval-ablation-sampling}
% \end{figure}

\paragraph{\texttt{Global-PS} selection strategy}
Figure~\ref{fig:eval-ablation-adaptive} demonstrates how our Global-PS planning strategy efficiently determines the best \texttt{global-ps} to achieve rapid convergence to a high accuracy. 
In comparison with a fixed, small \texttt{global-ps} value of 1, \sys delivers an accuracy boost of 43.6\%. 
And when pitted against a larger, fixed \texttt{global-ps} value of 50, \sys achieves equivalent accuracy of 85\% but 2.99$\times$ faster. 
This enhancement stems from that \sys dynamically adapts the \texttt{global-ps} value as training progresses. 
It starts with a smaller \texttt{global-ps} to rapidly gain basic accuracy and transitions to a larger \texttt{global-ps} to hone in on higher accuracy in the later stages.

\paragraph{Sensitivity of discriminative sampling}
Figure~\ref{fig:eval-ablation-sampling} illustrates how varying sampling ratios influence the convergence performance. 
A 20\% sampling ratio emerges as the most efficient, offering the best runtime improvement (2.33$\times$) compared to not employing any sampling (equivalent to a 100\% sampling ratio). 
Using a sampling ratio in the range of 10-40\% yields comparable runtime benefits, suggesting that there's flexibility in choosing an effective hyperparameter, ensuring robustness of the sampling method. 
However, opting for extremely low sampling ratios is not always beneficial. 
For instance, a mere 1\% sampling ratio hampers the convergence speed, taking 1.42$\times$ longer than the optimal 20\% ratio. 
This is due to the limited gradient update directions, resulting in a decelerated convergence.

    \section{Related Work}\label{sec:related}

\noindent \textbf{FedLLM}
With the recent rise of transformers and its  variants~\cite{vaswani2017attention,devlin2018bert,sanh2019distilbert, hou2020dynabert, liu2020fastbert}, LLMs have achieved great success in various domians, including CV, NLP, etc.
While its success is largely attributed to the pre-training paradigm, the privacy issue of LLMs has been a major concern, e.g., extraction attacks~\cite{deng2021tag, zhu2019deep, balunovic2022lamp}.
FedLLM is a promising direction to releave the privacy tension of LLMs~\cite{zhuang2023foundation}.
%\textbf{Parameter-efficient FedLLM}
%With the rise of GPT-3~\cite{brown2020language}, the pre-training size of LLMs has been increasing rapidly, which poses a huge challenge to the efficiency of LLMs~\cite{bommasani2021opportunities}.
To tackle the tight resource constraint on devices, most recent efforts resort to parameter-efficient fine-tuning (PEFT) techniques such as Adapter and LoRa~\cite{pfeiffer2020adapterhub,zaken2021bitfit,hu2021lora,he2021towards}.
For instance, FedAdapter~\cite{cai2022autofednlp} addresses the adapter configuration problem for FedLLM.
However, in $\S$\ref{sec:bkgnd} we show that PEFT-based FedLLM only mitigates the network bottleneck, but cannot significantly optimize the memory or speed of on-devce training.
\sys is built atop PEFT but significantly reduces the device-side overhead and accelerates convergence.
We refer to the latest resource-efficient LLM survey~\cite{xu2024survey} for more information.

\paragraph{Backpropagation-free training}
Backpropagation is the most widely adopted but not the only way to train neural networks, such as zero-order optimization proposed in early 80s~\cite{barto1987gradient}.
As models getting larger, reserachers are realizing that backpropagation becomes a burden to DNN training pipeline.
Thus, a few new paradigms for backpropagation-free training are proposed~\cite{hinton2022forward, ma2020hsic, baydin2022gradients, sun2022bbtv2}.
%For example, \cite{hinton2022forward} proposes the forward-forward algorithm to investigate new learning procedure for neural networks.
This work is built atop forward gradient for being mobile friendly.
However, prior literature~\cite{baydin2022gradients,ren2022scaling, park2023fedfwd} of forward gradient still use toy models (MLP or small CNNs).
BBTv2~\cite{sun2022bbtv2} applys gradient-free methods to optimize the LLM prompts in a central cloud.
%\cite{baydin2022gradients} proposes forward-gradient but only evaluated on a small dataset (MNIST) and a small model (MLP).
%However, very few of them focus on federated learning. 
BAFFLE~\cite{feng2023does} and FedZeN~\cite{maritan2023fedzen} are two concurrent work to \sys, which combine backpropagation-free training with FL.
However, it is not designed for LLM and lacks the two key techniques as presented in $\S$\ref{sec:design-adaptive} and $\S$\ref{sec:design-sampling} that make backpropagation-free FL more systematic and efficient.
Another concurrent work FedKSeed~\cite{qin2023federated} aims to enable full-parameter zeroth-order optimization of FedLLMs;
it reduces the network cost by updating a scalar gradient accumulator instead of forward gradients vectors.
However, it requires hundreds of seconds to decode the gradients from the scalar accumulator and excessive local training steps for each client.
%To our best knowledge, \sys is the very first system that leverages backpropagation-free training algorithm to enable cross-device FL of billion-size LLM like LLaMA-7B.

\noindent \textbf{FL Optimizations}
There have been tremendous efforts in making cross-device FL more efficient, including communication efficiency~\cite{bonawitz2019towards1,yang2021characterizing}, model compression/quantization~\cite{wu2018error, bernstein2018signsgd}, client/data sampling~\cite{li2021hermes,lipyramidfl, nishio2019client, xu2020client, wang2021device, lai2020oort, zhao2021quality, li2021sample}, and on-device training speedup~\cite{xu2022mandheling,wang2022melon}.
%To mitigate the heterogeneity of client devices (therefore stragglers), Abdelmoniem et al.~\cite{abdelmoniem2021towards} ask each client device to quantize their local model adaptively.
%Hermes~\cite{li2021hermes} guides different mobile clients to find a small subnetwork through structured pruning for local training.
%Another line of those work focus on intelligent client selection and data sampling~\cite{li2021hermes,lipyramidfl, nishio2019client, xu2020client, wang2021device, lai2020oort, zhao2021quality, li2021sample}.
%propose a fine-grained client selection framework for efficient federated learning.
%It exploits the data and system heterogeneity within akk clients to profile their utility more efficiently.
%As a result, more devices can participate in the future rounds.
%The accuracy of aggregated model is therefore improved. 
%Yang et al. characterizes FL in the face of real-world device traces, highlighting the significance of optimizing on-device training in FL ~\cite{yang2021characterizing}.
However, their benefits are modest compared to the gap between the LLMs like LLaMA and the resource constraint of mobile devices.
\sys innovates the FL protocol by abandoning backpropagation and therefore brings much more significant improvements.

    \section{Conclusions}\label{sec:conclusions}
In this study, we introduce \sys, the pioneering federated learning framework for LLMs that operates without backpropagation. 
It borrows the wisdom from the forward gradients method and applies it to FedLLM training, so as to avoid memory-intensive backpropagation and enable scalable training of LLMs on mobile devices.
To generate forward gradients more efficiently and precisely, we employ an adaptive perturbation generator that determines the number of perturbations on the fly. Additionally, we incorporate a discriminative sampler to selectively screen the generated perturbations. 
Comparative analyses reveal that \sys outperforms contemporary FedLLM methods in both convergence time and scalability.
% \textbf{Privacy aspect?}

% \textbf{Scalability aspect?}

    \balance
    \bibliographystyle{plain}
    \bibliography{bib/ref-mwx,bib/ref-cdq}

\end{document}